\documentclass[sigconf,nonacm]{aamas}

\usepackage{balance} 
\usepackage{subcaption}

\usepackage{multirow}

\setcopyright{ifaamas}
\acmConference[AAMAS '26]{Proc.\@ of the 25th International Conference
on Autonomous Agents and Multiagent Systems (AAMAS 2026)}{May 25 -- 29, 2026}
{Paphos, Cyprus}{C.~Amato, L.~Dennis, V.~Mascardi, J.~Thangarajah (eds.)}
\copyrightyear{2026}
\acmYear{2026}
\acmDOI{}
\acmPrice{}
\acmISBN{}

\acmSubmissionID{829}

\title[AAMAS-2026 Formatting Instructions]{Influencing LLM Multi-Agent Dialogue via Policy-Parameterized Prompts}

\author{Hongbo Bo}
\affiliation{
  \institution{University of Bristol}
  \city{Bristol}
  \country{United Kingdom}}
\email{hongbo.bo@bristol.ac.uk}

\author{Jingyu Hu}
\affiliation{
  \institution{University of Bristol}
  \city{Bristol}
  \country{United Kingdom}}
\email{jingyu.hu@bristol.ac.uk}

\author{Weiru Liu}
\affiliation{
  \institution{University of Bristol}
  \city{Bristol}
  \country{United Kingdom}}
\email{weiru.liu@bristol.ac.uk}

\begin{abstract}
Large Language Models (LLMs) have emerged as a new paradigm for multi-agent systems.
However, existing research on the behaviour of LLM-based multi-agents relies on ad hoc prompts and lacks a principled policy perspective. Different from reinforcement learning, we investigate whether prompt-as-action can be parameterized so as to construct a lightweight policy which consists of a sequence of state-action pairs to influence conversational behaviours without training. Our framework regards prompts as actions executed by LLMs, and dynamically constructs prompts through five components based on the current state of the agent. To test the effectiveness of parameterized control, we evaluated the dialogue flow based on five indicators: responsiveness, rebuttal, evidence usage, non-repetition, and stance shift. We conduct experiments using different LLM-driven agents in two discussion scenarios related to the general public and show that prompt parameterization can influence the dialogue dynamics. 
This result shows that policy-parameterised prompts offer a simple and effective mechanism to influence the dialogue process, which will help the research of multi-agent systems in the direction of social simulation.
\end{abstract}


\newcommand{\BibTeX}{\rm B\kern-.05em{\sc i\kern-.025em b}\kern-.08em\TeX}

\begin{document}


\pagestyle{fancy}
\fancyhead{}


\maketitle 
\footnotetext{Accepted at AAMAS 2026}


\section{Introduction}

Traditional multi-agent simulations often rely on explicit modelling or reinforcement learning to train policies~\cite{gronauer2022multi,zhu2024survey,wu2025multi}, enabling agents to learn how to respond and interact. In contrast, Large Language Models (LLMs) have emerged as a new paradigm for multi-agent systems \cite{li2024survey,guo2024large}, where agents inherently possess natural language generation and knowledge retrieval capabilities without requiring additional training for basic interaction. Based on this potential, LLM-based multi-agent systems have been widely used in social simulation tasks by assigning roles, tasks, and instructions to LLMs in recent studies~\cite{park2023generative,gao2023s3}.

LLMs agents in these approaches typically communicate with each other using ad hoc prompts, focusing on whether agents can align human preferences and behaviors like collaboration \cite{hong2024metagpt,yang2024swe,wu2023autogen}, negotiation \cite{estornell2024multi,abdelnabi2023llm,liang2023encouraging} and provide rational plans \cite{fan2024can,hao2023reasoning}. However, these methods lack a principled framework for treating communication strategies as policies to systematically control agent behaviors. 
Without sufficient attention to how agent dialogue can be deliberately shaped and optimized, it becomes difficult to predict agent behavior, optimize communication patterns, and transfer insights across different tasks.

To address this limitation, we expect to propose a principled way to conceptualize and operationalize agent communication strategies, one that allows us to formally define, compare, and optimize different strategies for multi-agent dialogue. Specifically, this study focuses on how policy-parameterised prompts can be utilised to influence conversational behaviours in multi-agent discussion dialogues.

This work considers the input prompt itself as an action generated by a lightweight form of policy parameterisation. Specifically, we decompose the prompt into five components: task and persona description (T), dialogue history memory (M), external knowledge base (D), rule template (R), and weight (W). By adaptively parameterising these components to allow different levels of influence on LLM agents, we can directly impose the significance of different factors on the utterance style of dialogue, thereby modulating their conversational behaviours without any additional training.

Within this policy framework, we implement multi-role-based agents with distinct stances and knowledge bases, and engage them in multi-round dialogue on issues related to the general public. To further enhance adaptivity, we design an adaptive weight scheduler that automatically adjusts the reliance on T/M/D during the dialogue, based on temporal trends and behavioural feedback. To quantify the effects of different control strategies, we propose a set of evaluation metrics, including responsiveness, rebuttal, non-repetition, evidence usage, and stance shift. These metrics allow us to systematically compare dialogue differences under various prompt control modes and to observe the evolution of group stances over time.

Our designed framework is to answer the following research questions. RQ1: Can prompt control be used as a lightweight form of policy parameterisation to regulate the conversational behaviours of LLM-based multi-agent systems? RQ2: Do different prompt control strategies (e.g., rule templates and weight scheduling) lead to significant behavioural differences, such as changes in responsiveness, evidence usage, and stance evolution? To address these questions, we design two discussion scenarios related to the general public land resources use and educational resource allocation, each involving three agents with distinct personas driven by different LLMs. In each scenario, the agents interact over multiple rounds of dialogue, with prompts dynamically constructed from task, memory, and evidence components. We systematically vary control strategies by enabling or disabling rule templates and adjusting weight parameters, and evaluate the resulting dialogues along five proposed evaluation metrics.  Our results show that prompt parameterisation can effectively influence dialogue, with different strategies leading to distinct patterns in rebuttal, evidence usage, and stance shift.

Overall, the core of this study lies in treating prompt control as a lightweight policy to achieve structured regulation of LLM multi-agent dialogue, thereby exploring a social simulation pathway that is distinct from traditional training-based approaches.

\section{Related Work}

\paragraph{LLMs Agents and Social Simulation.}

Recent works have explored the use of LLMs as agents in social simulation, aiming to explore realistic social phenomena by simulating social behaviours. 
\cite{park2023generative} applied LLMs with memory to build generative agents, and found agents exhibit emergent social behaviors such as information diffusion, relationship formation, and coordinating attendance in a sandbox environment. Similar works~\cite{lin2023agentsims, fan2022minedojo,liu2023training} also utilized LLMs to build the sandbox environment to study social behaviors. \cite{park2022social} developed a simulated community of 1,000 personas by design prompt chains to generate behaviors like posting, replying, and also anti-social actions.
$S^3$ system \cite{gao2023s3} uses LLM agents to simulate users' emotions, attitudes, and interaction behaviours in the social networks, by endowing the agents in the system with the ability to perceive the informational environment. CAMEL \cite{li2023camel} is a multi-agent role-playing framework proposed to leverage inception prompting to initialise roles, tasks, and dialogue formats, enabling LLM-based chat agents to collaborate on complex tasks. \cite{ashery2025emergent} applies multi-agent LLMs to simulate agent populations, and their results show that such populations can develop social conventions and collective biases through decentralized interactions. \cite{piao2025agentsociety} extends society simulations to large scale with over 10k agents and 5 million interactions and analyzed their believable individual and emergent social behaviors.
Other studies \cite{hong2024metagpt, liang2023encouraging} validated the effectiveness of multi-agent systems in collaboration and debate tasks.
However, prompts in these systems are typically ad hoc, without a principled treatment as policies.

\paragraph{LLM Agents Formalization.}
Several works have sought to abstract LLM-based agents into decision-theoretic frameworks, drawing inspiration from reinforcement learning (RL) or the Belief-Desire-Intention (BDI) model. For instance, LLMs have been cast as policies mapping states to natural-language actions, and also act as a world model combined with Monte Carlo Tree Search for planning search~\cite{hao2023reasoning}.  BDIPrompting \cite{jang2023structured} integrates the BDI model into prompt design to improve proactive action planning and transparency in LLMs.
Reflexion\cite{shinn2023reflexion} is a framework that reinforces LLM agents through verbal feedback, parameterizing a policy as memory to enable trial-and-error learning via self-reflection. \cite{fan2024can} study whether LLMs can serve as rational players in game theory, by providing inputs with preferences and rules to build belief and desire and then plan the optimal action.  \cite{xu2023language} combined LLMs with reinforcement learning (RL) to enable agents to learn strategic language communication and decision-making for the Werewolf game.
These studies focus on questions—whether LLMs can act rationally or align with human decisions, but do not address how LLMs can be controlled in multi-agent. Also, in their study, LLMs were used to generate decisions rather than to execute actions.

\section{Prompt Control as Lightweight Policy Parameterization}
\label{sec:method}

\begin{figure*}
    \centering
    
    \includegraphics[width=\linewidth]{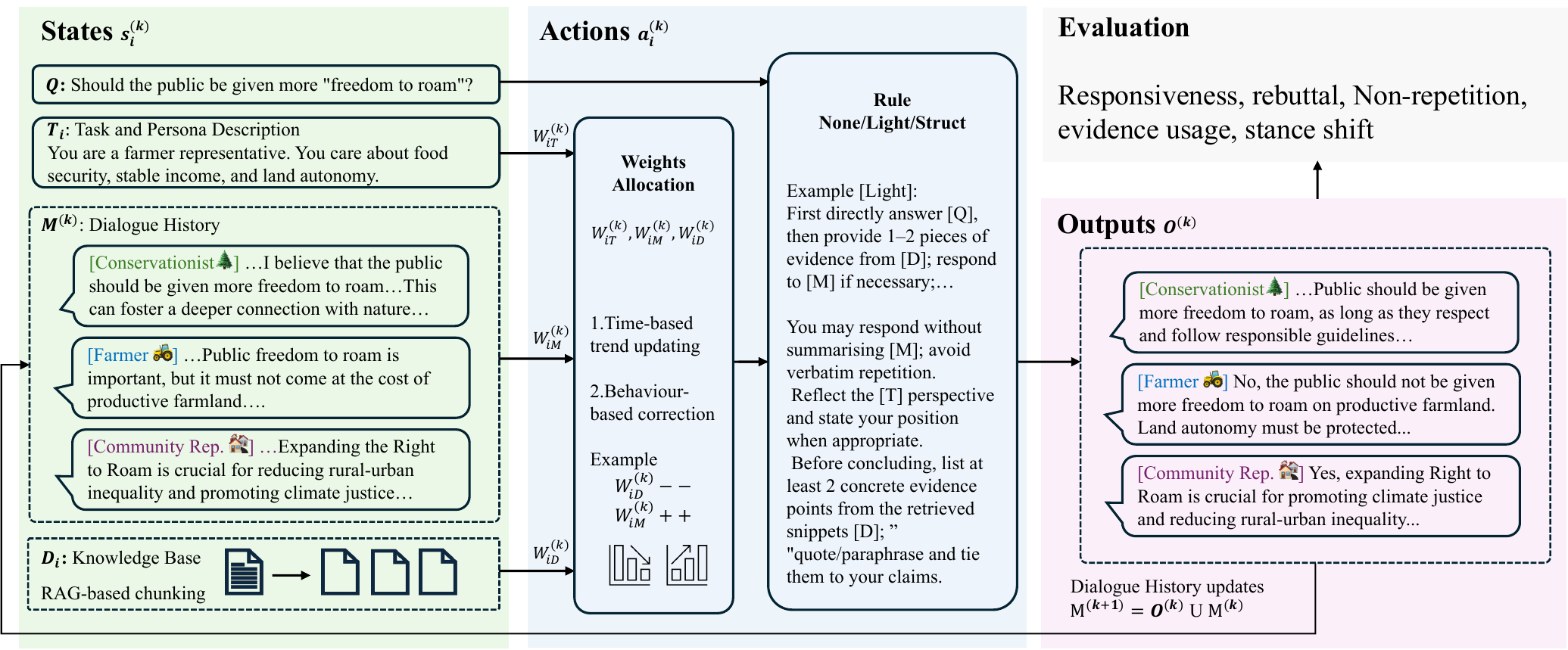}
    \caption{The overall framework illustrates the process from an agent’s state representation to action generation, LLM-based action execution, and evaluation.}
    \label{fig:prompt}
\end{figure*}

This section describes how our proposed method influences LLM multi-agent dialogue via policy-parameterized prompts. 
Figure~\ref{fig:prompt} illustrates the overall framework of our method.

\subsection{Multi-Agent Formalization}

We formulate multi-agent discussions as a controllable state--action process where policies are specified directly through prompt construction.  
Specifically, LLM multi-agent conversation is formalized as a prompt-parameterized process: We define $N$ agents engaging in $K$ rounds of discussion.
For each conversation round $k \in \{1,2,..., K\}$, 
each agent $\mathcal{A}_i$ has a corresponding state $s_i^{(k)}$ composed of agent memory, evidence and task settings. The policy $\pi_i$ then maps the state to a constructed prompt as agent's action $a_i^{(k)}$.
We achieve parameterized control over agent behavior through prompts via rules template $R$ and weights vector $W$ adjustment.

\subsubsection{Agent}
Each agent is represented as a quadruple
$\mathcal{A}_i = \{Q,T_i,$ 
$ D_i, LLM_i\}$,
where $Q$ is the global discussion query among all agents provided by the user,  $T_i$ includes the task description and agent $\mathcal{A}_i$ persona setting, 
$D_i$ is the agent $\mathcal{A}_i$ complete role-specific knowledge data, $LLM_i$ is the executor of actions to generate dialogues by the agent $\mathcal{A}_i$. As this study specifically examines LLM-based agents, we consider the LLM as part of the agent's capability that it can call to execute actions (similar to calling other functions, etc.).

\subsubsection{State and Retrieval}
\label{sec:retrieval}
At round $k$, the dialogue memory is denoted as $M^{(k)}$, consisting of all $k$ rounds of utterances. 
From the full knowledge base $D_i$, the agent retrieves a subset $\hat D_i^{(k)}$ via embedding similarity search:
\[
\hat D_i^{(k)} =
\begin{cases}
\operatorname*{Top\text{-}n}_{c \in \mathcal{C}(D_i)} \cos(\phi(c), \phi(Q)), & k=1, \\[6pt]
\operatorname*{Top\text{-}n}_{c \in \mathcal{C}(D_i)} \cos(\phi(c), \phi(M^{(k)})), & k > 1,
\end{cases}
\]
where $\mathcal{C}(D_i)$ is the set of chunks obtained by segmentation and $\phi(\cdot)$ is the embedding function used in Retrieval-Augmented Generation (RAG)~\cite{lewis2020retrieval}.

The \emph{state} of agent $\mathcal{A}_i$ is composed of $s_i^{(k)} = \{T_i,\, Q,\, \hat M^{(k)},\, \hat D_i^{(k)}\}$,
where $\hat M^{(k)}\subseteq M^{(k)}$ denotes the subset of dialogue memory extracted for the current round, consisting of the recent dialogues, while $\hat M^{(1)}$ is empty at initialisation. In this study, we adopt a \emph{shared message pool} setting~\cite{guo2024large}, in which all agents have access to a common dialogue history; thus, the memory $M^{(k)}$ is shared rather than individual-based. Our method can also be extended to alternative communication configurations, such as layered, decentralised or centralised.

\subsubsection{Prompt-as-Action}

We adopt the prompt-as-action view: the policy $\pi_i$ maps the state $s_i^{(k)}$ to an action $a_i^{(k)}$, where the action is the constructed prompt: $a_i^{(k)} = \pi_i(s_i^{(k)})$.
Here, the policy $\pi_i$ generates a action $a_i^{(k)}$   based on $<R\oplus W_i^{(k)}>$ under state $s_i$ , where $R$ is the optional rule template and $W_i^{(k)}=\langle w_{iT}^{(k)},w_{iM}^{(k)},w_{iD}^{(k)} \rangle$ are weights controlling how strongly persona, memory, or retrieved knowledge are emphasized. 
The operator $\oplus$ denotes the coupling of rules and weights into a single prompt specification. Concretely, $\oplus$ yields (a) tiered micro-instructions attached in prompts next to the affected [T]/[M]/[D] blocks when a weight is low/high (in Section~\ref{sec:weight}), and (b) an optional rule skeleton inserted in the [R] block when enabled (in Section~\ref{sec:rule}). 
In this way, policy-parameterised prompts provide a lightweight mechanism to steer the dialogue process of LLM-based agents. The policy in this study will produce a series of state-action pairs and this can be treated as a trajectory in RL policy learning, noting that in our framework, the trajectory length is determined by the number of rounds $K$ of dialogues required.

\subsubsection{Action Execution and Output}
The action $a_i^{(k)}$ (the prompt) is executed by the LLM, yielding a concrete natural-language output 
$o_i^{(k)} = \mathrm{LLM_i}(a_i^{(k)})$.
In each round $k$, all agents follow the same action generation to generate their own actions for producing their respective output $o_i^{(k)}$. These outputs are then aggregated and appended to $M^{(k)}$ to form the global dialogue memory $M^{(k+1)}$ for the next round, which in turn influences subsequent retrieval $M^{(k+1)} = \{\cup_{i=1}^{N} o_i^{(k)}\} \cup M^{(k)}$.

\subsection{Policy Parameterization}

\subsubsection{Rule Templates R}
\label{sec:rule}
Besides the four information sources $Q$, $T$, $M$, and $D$, we introduce an optional component $R$ (rule template), which is designed to steer the interaction behaviour of the agent in a controllable way. The use of $R$ allows the agent to specify the format of the output and the way the information is used, thereby shaping its conversational behaviour without changing the underlying model parameters. We design three rule templates that represent incremental degrees of structural constraint:

\begin{itemize}
    \item \textit{None}: No explicit structural instruction is given; the agent directly generates a response based on the concatenated information blocks \([T], [M], [D], [Q]\), without any additional ordering or format control. 
    \item \textit{Light}: Provides minimal structure by specifying a basic response order and length constraint. 
    `First directly answer [Q], then provide 1–2 pieces of evidence from [D]; respond to [M] if necessary; limit the response to at most {N} sentences.'
    \item \textit{Struct}: 
    Enforces a detailed reasoning structure by decomposing the discussion into specific categories of key points (supporting, opposing, conflicting, cooperative).  `First extract four types of key points in order (no more than 3 each) from [M]:  
    1) arguments supporting the goal, 2) arguments threatening the goal,  
    3) unresolved points of conflict, 4) potential opportunities for cooperation;  
    then generate a response of no more than 3 sentences based on these points, giving priority to citing [D].'
\end{itemize}


This allows us to have a lightweight policy that dynamically generates prompts (or actions) as the dialogue continues so as to steer the emergent discussion patterns among agents.

\subsubsection{Weights Design.}
\label{sec:weight}

To better refine the policy impact on influencing agent behaviors, we define a weight set $W_i = \{ w_{iT}^{(k)}, w_{iM}^{(k)},$ $ w_{iD}^{(k)} \}$ for each agent $\mathcal{A}_i$. Each weight $w \in W_i$ takes values in $[0,2]$ and can dynamically adjust $\mathcal{A}_i$'s reliance on the corresponding component (T, M, and D) during conversations. Each weight $w$ is then mapped to a three-tier signal system: \textit{low} if $w \in[0, 0.85]$; \textit{mid} if $w \in [0.85,1.25]$; \textit{high} if $w \in [1.25,2]$. Each tier corresponds to specific behavioral instructions for the three components. The simplified instructions for each component and tier are outlined below,
and complete instructions are provided in Appendix~\ref{app:instructions}.

\begin{itemize}
    \item T:
    \begin{itemize}
        \item low: keep persona implicit; focus on arguments;
        \item mid: reflect the assigned role’s perspective and express stance when relevant;
        \item high: explicitly state role/stance first, then justify;
    \end{itemize}
    \item M:
    \begin{itemize}
        \item low: may reply without summarising; avoid repetition;
        \item mid: consider prior discussion to maintain contextual coherence;
        \item high: begin with a brief recap and resolve pending points;
    \end{itemize}
    \item D:
    \begin{itemize}
        \item low: may proceed without citing retrieved snippets if not essential;
        \item mid: Use retrieved snippets to support key claims when appropriate;
        \item high: provide concrete evidence items before concluding;
    \end{itemize}
\end{itemize}

\subsubsection{Adaptive Weights}
To better reflect the evolving nature of dialogue, we introduce time-based trend updating and behaviour-based correction to update the prompt weights.

\paragraph{Time-based trend updating}
We assume agents should rely more on $D$ at early rounds to establish their stance, and rely more on the dialogue history $M$ in later rounds to engage with the ongoing debate. Accordingly, we define  weights as:
\[
\begin{aligned}
w_{iM}^{(k)} &= \min\{w_{iM}^{(0)} + 0.1k,\; 2.0\},\\
w_{iD}^{(k)} &= \max\{w_{iD}^{(0)} - 0.1k,\; 0.5\},\\
w_{iT}^{(k)} &\equiv w_{iT}^{(0)}.
\end{aligned}
\]

\paragraph{Behaviour-based correction}
After observing the agent’s previous response, we apply an update operator $\min(\cdot)$ parameterized by $\alpha$ (different update operators can also be defined). Specifically:
\begin{itemize}
    \item If agent failed to use D in previous round $k-1$:  \[\quad w_{iD}^{(k)} \gets \min\{w_{iD}^{(k)}+\alpha,\;2.0\}\]
    \item If agent failed to respond to M: \[\quad w_{iM}^{(k)} \gets \min\{w_{iM}^{(k)}+\alpha,\;2.0\}\]
\end{itemize}

\subsection{Evaluation}

We evaluate the effectiveness of policy parameterization by five proposed evaluation metrics on the output $o$ executed by the agent over the state--action pairs $(s,a)$.  We conduct the evaluation using an LLM as the judge model and a text embedding model as the embedding backend.
Each metric $m(o_i^{(k)})$ captures a distinct behavioral dimension:

\begin{itemize}
    \item \textbf{Responsiveness (Resp.)}  $m_{\text{resp}}o_i^{(k)}$: 
    Returns $1$ if $o_i^{(k)}$ addresses the most recent utterance contained in $M^{(k)}$, and $0$ otherwise. 
    The judge model is prompted with the previous and current utterances to produce a binary decision.

    \item \textbf{Rebuttal} $m_{\text{rebut}}o_i^{(k)}$: 
    Returns $1$ if $o_i^{(k)}$ explicitly \textit{opposes} the most recent utterance in $M^{(k)}$, as classified by the judge model; $0$ otherwise.  This directly captures  `whether a rebuttal occurs.'

    \item \textbf{Non-repetition (Non-rep.)} $m_{\text{nrep}}o_i^{(k)}$: 
    Measures novelty of $o_i^{(k)}$ with respect to the agent’s own previous action $o_i^{(k-1)}$,  defined as one minus the similarity between the current and the agent’s previous utterance. Similarity is defined as the maximum of string overlaps $over()$ and embedding cosine similarity $cos()$, with additional penalties for repeated sentence openings. Defined as
    \[
    m_{\text{nrep}}o_i^{(k)} = 1 - \max\{over(o_i^{(k)},o_i^{(k-1)}), cos(o_i^{(k)},o_i^{(k-1)})\}
    \]
    
    \item \textbf{Evidence usage (Evid.)} $m_{\text{evid}}o_i^{(k)}$: 
    Returns $1$ if key phrases from the retrieved knowledge component $D_i$ appear in $o_i^{(k)}$, and $0$ otherwise.
    
    \item \textbf{Stance shift} $m_{\text{stance}}o_i^{(k)}$: 
    Computes the cosine similarity between the embedding of $o_i^{(k)}$ and the embedding of the persona description $T_i$. 
    Tracking $m_{\text{stance}}$ across rounds reveals whether the agent remains aligned with or diverges from its original stance.
\end{itemize}

Each metric is averaged over all dialogue rounds for a given agent, and the overall performance is obtained by taking the mean across all agents.

\begin{table*}[t]
\centering
\caption{Overall performance of policies under different rule templates on 5 queries over 10-round conversation. Values are the mean $\pm$ std of 5 runs experiments. Because Rebuttal and Evid. are binary (0/1), the mean is just the proportion of 1s and the standard deviation can appear large relative to the mean.}
\label{tab:all_metrics}
\begin{tabular}{c|c|ccccc|ccccc}
\toprule
&& \multicolumn{5}{c|}{\textbf{Land}} & \multicolumn{5}{c}{\textbf{Education}} \\\midrule

\textbf{Query} & \textbf{Rule} & \textbf{Resp.} & \textbf{Rebuttal} & \textbf{Non-rep.} & \textbf{Evid.} & \textbf{Stance} & \textbf{Resp.} & \textbf{Rebuttal} & \textbf{Non-rep.} & \textbf{Evid.} & \textbf{Stance}\\
\midrule
Q1 & None & $0.88_{\pm 0.33}$ & $0.19_{\pm 0.40}$ & $0.47_{\pm 0.42}$ & $0.10_{\pm 0.30}$ & $0.52_{\pm 0.08}$ & $0.88_{\pm 0.33}$ & $0.25_{\pm 0.43}$ & $0.45_{\pm 0.41}$ & $0.13_{\pm 0.34}$ & $0.52_{\pm 0.09}$ \\
 & Light & $0.90_{\pm 0.30}$ & $0.21_{\pm 0.41}$ & $0.33_{\pm 0.35}$ & $0.32_{\pm 0.47}$ & $0.51_{\pm 0.09}$ & $0.87_{\pm 0.34}$ & $0.27_{\pm 0.44}$ & $0.41_{\pm 0.36}$ & $0.25_{\pm 0.44}$ & $0.52_{\pm 0.11}$ \\
 & Struct & $0.74_{\pm 0.44}$ & $0.27_{\pm 0.45}$ & $0.58_{\pm 0.39}$ & $0.31_{\pm 0.46}$ & $0.49_{\pm 0.09}$ & $0.89_{\pm 0.32}$ & $0.09_{\pm 0.29}$ & $0.54_{\pm 0.37}$ & $0.17_{\pm 0.37}$ & $0.52_{\pm 0.11}$ \\
\midrule
Q2 & None & $0.87_{\pm 0.34}$ & $0.37_{\pm 0.48}$ & $0.47_{\pm 0.44}$ & $0.11_{\pm 0.31}$ & $0.49_{\pm 0.08}$ & $0.85_{\pm 0.35}$ & $0.15_{\pm 0.36}$ & $0.41_{\pm 0.41}$ & $0.37_{\pm 0.48}$ & $0.48_{\pm 0.07}$ \\
 & Light & $0.89_{\pm 0.31}$ & $0.31_{\pm 0.47}$ & $0.32_{\pm 0.34}$ & $0.27_{\pm 0.44}$ & $0.45_{\pm 0.06}$ & $0.89_{\pm 0.32}$ & $0.06_{\pm 0.24}$ & $0.35_{\pm 0.35}$ & $0.41_{\pm 0.49}$ & $0.45_{\pm 0.08}$ \\
 & Struct & $0.79_{\pm 0.41}$ & $0.17_{\pm 0.38}$ & $0.64_{\pm 0.34}$ & $0.05_{\pm 0.21}$ & $0.46_{\pm 0.07}$ & $0.90_{\pm 0.30}$ & $0.06_{\pm 0.24}$ & $0.68_{\pm 0.30}$ & $0.21_{\pm 0.41}$ & $0.52_{\pm 0.07}$ \\
\midrule
Q3 & None & $0.90_{\pm 0.30}$ & $0.00_{\pm 0.00}$ & $0.42_{\pm 0.42}$ & $0.09_{\pm 0.28}$ & $0.47_{\pm 0.07}$ & $0.88_{\pm 0.33}$ & $0.17_{\pm 0.38}$ & $0.47_{\pm 0.39}$ & $0.10_{\pm 0.30}$ & $0.49_{\pm 0.06}$ \\
 & Light & $0.90_{\pm 0.30}$ & $0.08_{\pm 0.27}$ & $0.38_{\pm 0.38}$ & $0.19_{\pm 0.39}$ & $0.45_{\pm 0.05}$ & $0.89_{\pm 0.31}$ & $0.15_{\pm 0.36}$ & $0.36_{\pm 0.35}$ & $0.25_{\pm 0.43}$ & $0.49_{\pm 0.05}$ \\
 & Struct & $0.86_{\pm 0.35}$ & $0.04_{\pm 0.20}$ & $0.70_{\pm 0.32}$ & $0.15_{\pm 0.35}$ & $0.45_{\pm 0.07}$ & $0.89_{\pm 0.32}$ & $0.06_{\pm 0.24}$ & $0.61_{\pm 0.32}$ & $0.24_{\pm 0.43}$ & $0.51_{\pm 0.05}$ \\
\midrule
Q4 & None & $0.79_{\pm 0.41}$ & $0.43_{\pm 0.50}$ & $0.34_{\pm 0.38}$ & $0.10_{\pm 0.30}$ & $0.49_{\pm 0.06}$ & $0.85_{\pm 0.35}$ & $0.13_{\pm 0.34}$ & $0.53_{\pm 0.35}$ & $0.10_{\pm 0.30}$ & $0.47_{\pm 0.11}$ \\
 & Light & $0.68_{\pm 0.47}$ & $0.51_{\pm 0.50}$ & $0.29_{\pm 0.35}$ & $0.50_{\pm 0.50}$ & $0.50_{\pm 0.08}$ & $0.88_{\pm 0.33}$ & $0.17_{\pm 0.37}$ & $0.39_{\pm 0.33}$ & $0.35_{\pm 0.48}$ & $0.45_{\pm 0.12}$ \\
 & Struct & $0.76_{\pm 0.43}$ & $0.37_{\pm 0.48}$ & $0.54_{\pm 0.36}$ & $0.28_{\pm 0.45}$ & $0.49_{\pm 0.07}$ & $0.88_{\pm 0.33}$ & $0.03_{\pm 0.18}$ & $0.65_{\pm 0.28}$ & $0.17_{\pm 0.38}$ & $0.49_{\pm 0.11}$ \\
\midrule
Q5 & None & $0.83_{\pm 0.37}$ & $0.40_{\pm 0.49}$ & $0.46_{\pm 0.41}$ & $0.08_{\pm 0.27}$ & $0.44_{\pm 0.07}$ & $0.78_{\pm 0.42}$ & $0.07_{\pm 0.25}$ & $0.48_{\pm 0.40}$ & $0.09_{\pm 0.28}$ & $0.44_{\pm 0.07}$ \\
 & Light & $0.87_{\pm 0.34}$ & $0.43_{\pm 0.50}$ & $0.32_{\pm 0.35}$ & $0.13_{\pm 0.33}$ & $0.44_{\pm 0.04}$ & $0.85_{\pm 0.36}$ & $0.00_{\pm 0.00}$ & $0.39_{\pm 0.36}$ & $0.27_{\pm 0.44}$ & $0.42_{\pm 0.05}$ \\
 & Struct & $0.87_{\pm 0.34}$ & $0.23_{\pm 0.42}$ & $0.68_{\pm 0.35}$ & $0.20_{\pm 0.40}$ & $0.47_{\pm 0.07}$ & $0.89_{\pm 0.31}$ & $0.09_{\pm 0.28}$ & $0.61_{\pm 0.34}$ & $0.22_{\pm 0.42}$ & $0.50_{\pm 0.07}$ \\
\midrule
Overall & None & $0.85_{\pm 0.35}$ & $0.28_{\pm 0.45}$ & $0.43_{\pm 0.42}$ & $0.10_{\pm 0.29}$ & $0.48_{\pm 0.07}$ &$0.85_{\pm 0.36}$ & $0.15_{\pm 0.36}$ & $0.47_{\pm 0.39}$ & $0.16_{\pm 0.37}$ & $0.48_{\pm 0.09}$  \\
 & Light & $0.85_{\pm 0.36}$ & $0.31_{\pm 0.46}$ & $0.33_{\pm 0.36}$ & $0.28_{\pm 0.45}$ & $0.47_{\pm 0.07}$ & $0.88_{\pm 0.33}$ & $0.13_{\pm 0.34}$ & $0.38_{\pm 0.35}$ & $0.30_{\pm 0.46}$ & $0.47_{\pm 0.09}$\\
 & Struct & $0.80_{\pm 0.40}$ & $0.22_{\pm 0.41}$ & $0.62_{\pm 0.36}$ & $0.20_{\pm 0.40}$ & $0.47_{\pm 0.08}$ & $0.89_{\pm 0.31}$ & $0.07_{\pm 0.25}$ & $0.62_{\pm 0.33}$ & $0.20_{\pm 0.40}$ & $0.51_{\pm 0.09}$ \\
\bottomrule

\end{tabular}
\end{table*}

\section{Experiments}
This section presents the experimental setup and evaluation of our proposed method.
The implementation code is available in the supplementary materials.

\subsection{Experiment Settings}
To evaluate the effectiveness of the proposed policy parameterization governing agent behaviour through experiments, we design two scenarios: \textit{Land Resource Use} (Land) and \textit{Educational Resource Allocation} (Education), and instantiate three agents with distinct personas with tasks (T) and knowledge base (D) for each scenario. The knowledge bases are collected from publicly available materials, including government policy documents, blogs, and relevant websites, and are further summarised and supplemented using ChatGPT-5\footnote{https://openai.com/index/introducing-gpt-5/}~\cite{openai2025chatgpt} to ensure consistency and clarity. While processing these materials, ChatGPT-5 simultaneously identifies key stakeholder perspectives and formulates them into corresponding agent roles and task descriptions (T). 
More information about these knowledge bases are provided in the appendix~\ref{app:materials} 
The three agents in each scenario are driven by three different LLMs, Qwen3-8B\footnote{Qwen3-8B https://qwen.ai/research}~\cite{yang2025qwen3}, Llama3-8B\footnote{Meta-Llama-3-8B https://www.llama.com/models/llama-3/}~\cite{dubey2024llama}, and Mistral-7B\footnote{Mistral 7B https://mistral.ai/news/announcing-mistral-7b}~\cite{jiang2023mistral7b}, which are shown in Table~\ref{tab:agents}.
\begin{table}[H]
\centering
\caption{Agents and LLM assignments across two scenarios.}
\label{tab:agents}
\begin{tabular}{ccccc}
\hline
\multicolumn{2}{c}{\textbf{Land}} & & \multicolumn{2}{c}{\textbf{Education}} \\
\cline{1-2} \cline{4-5}
\textbf{Agent} & \textbf{LLM} &  & \textbf{Agent} & \textbf{LLM} \\
\hline
Farmer        & Qwen3   & & Rural Teacher  & Qwen3 \\
Conservationist  & Llama3  &  & Urban Parent   & Llama3 \\
Community Rep.     & Mistral & & Policy Maker   & Mistral \\
\hline
\end{tabular}
\end{table}

Each agent is paired with its own role-specific external knowledge base and engages in 10 rounds of dialogue on a controversial topic query $Q$ (e.g., “Should farmland be converted to forest?”).
For retrieved knowledge ($\hat D_i^{(k)}$), we adopt a retrieval-augmented generation (RAG) setup: at each round, the agent retrieves the top-3 relevant passages from its knowledge base using the current topic and recent dialogue memory (following Subsection~\ref{sec:retrieval}), and appends them to the prompt.
Prompts are dynamically constructed from the framework introduced in Section~\ref{sec:method}.
We compare different prompt control strategies by varying the rule template $R$ while keeping weight parameters fixed to $w_T=1.0, w_M=1.0, w_D=1.0$. For evaluation, we use Llama3 as the judge model to classify each dialogue round along evaluation metrics, and the embedding model  all-MiniLM-L6-v2\footnote{\url{https://huggingface.co/sentence-transformers/all-MiniLM-L6-v2}} \cite{wang2020minilm} is employed to compute semantic similarities for the evaluation metrics. 
Each value of metrics represents the average score across all agents and all rounds. 
For each scenario, we conduct five independent runs per query over five public topic queries corresponding to the scenario.
Here are 5 topic queries for each scenario, which are manually formulated based on the content of the materials covering multiple aspects:
\begin{itemize}
\item Land:
\begin{itemize}
  \item  Q1: `Should the public be given more freedom to roam?'
  \item Q2: `Should farmers be restricted from expanding farmland in sensitive ecological areas?'
  \item Q3: `What responsibilities do farmers bear in addressing climate change?'
  \item Q4: `How should land planning in the UK be formulated over the next 50 years?'
  \item Q5: `Should public greenways be built on private farmland?'
\end{itemize}
\item Education:
\begin{itemize}
  \item Q1: `How should governments allocate limited education funds between rural schools with poor facilities and urban schools facing intense competition?'
  \item Q2: `Should reliance on standardized exams be reduced, given the disadvantages for rural students and the heavy pressure on urban students?'
  \item  Q3: `Should public universities adjust tuition policies to improve access for low-income rural students while maintaining fairness for urban families?'
  \item  Q4: `Should governments invest in digital infrastructure and AI tutors in rural schools before expanding them in urban schools?'
  \item  Q5: `Should education policy prioritize funding for student support services (e.g., boarding schools in rural areas, mental health in urban schools)?'
\end{itemize}
\end{itemize}

\begin{figure*}[t]
\centering
\begin{subfigure}[t]{0.32\textwidth}
    \centering
    \includegraphics[width=\linewidth]{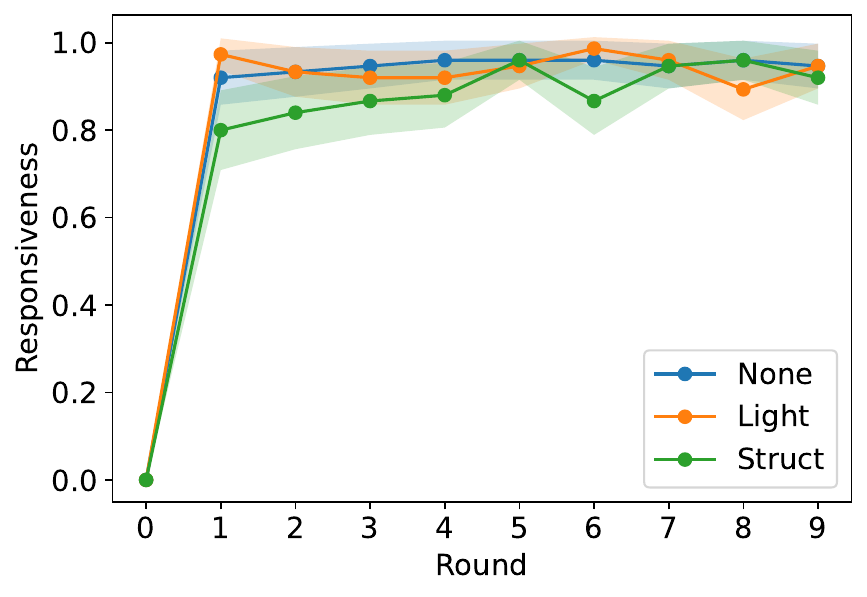}
    \caption{Responsiveness--Land}

\end{subfigure}
\hfill
\begin{subfigure}[t]{0.32\textwidth}
    \centering
\includegraphics[width=\linewidth]{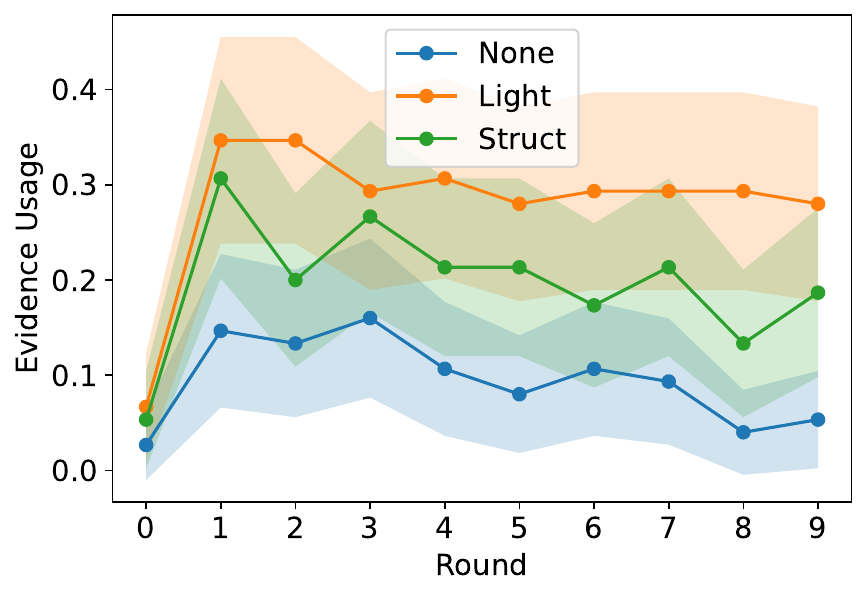}
    \caption{Evidence usage--Land}

\end{subfigure}
\hfill
\begin{subfigure}[t]{0.32\textwidth}
    \centering
    \includegraphics[width=\linewidth]{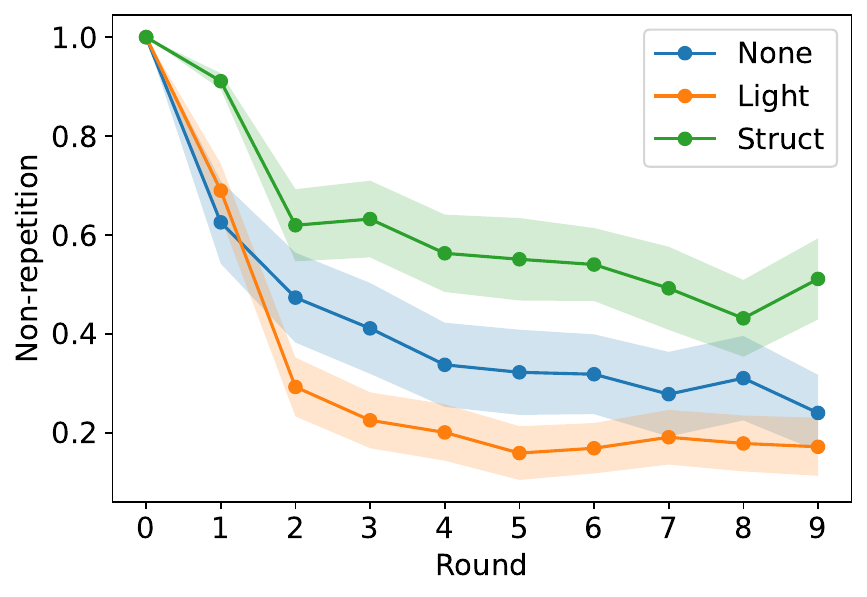}
    \caption{Non-repetition--Land}

\end{subfigure}
\begin{subfigure}[t]{0.32\textwidth}
    \centering
    \includegraphics[width=\linewidth]{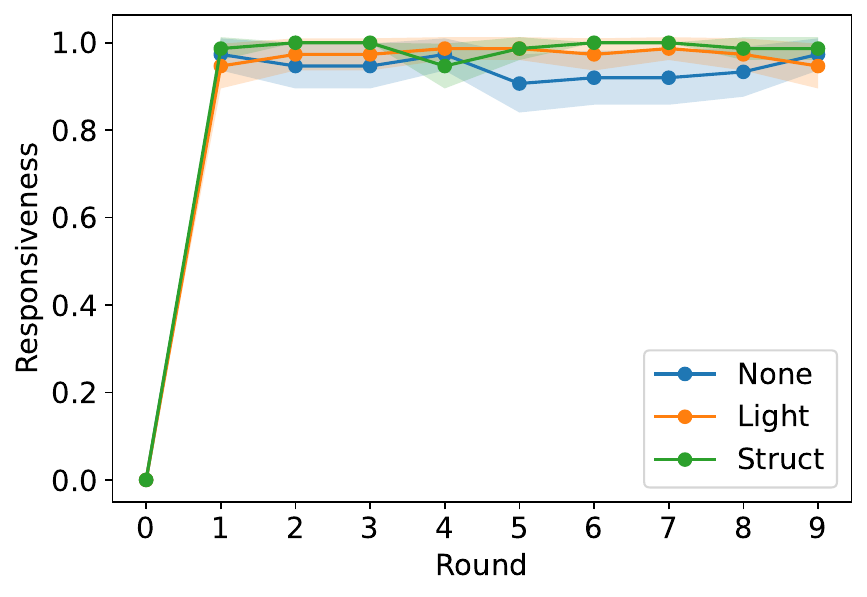}
    \caption{Responsiveness--Education}

\end{subfigure}
\hfill
\begin{subfigure}[t]{0.32\textwidth}
    \centering
\includegraphics[width=\linewidth]{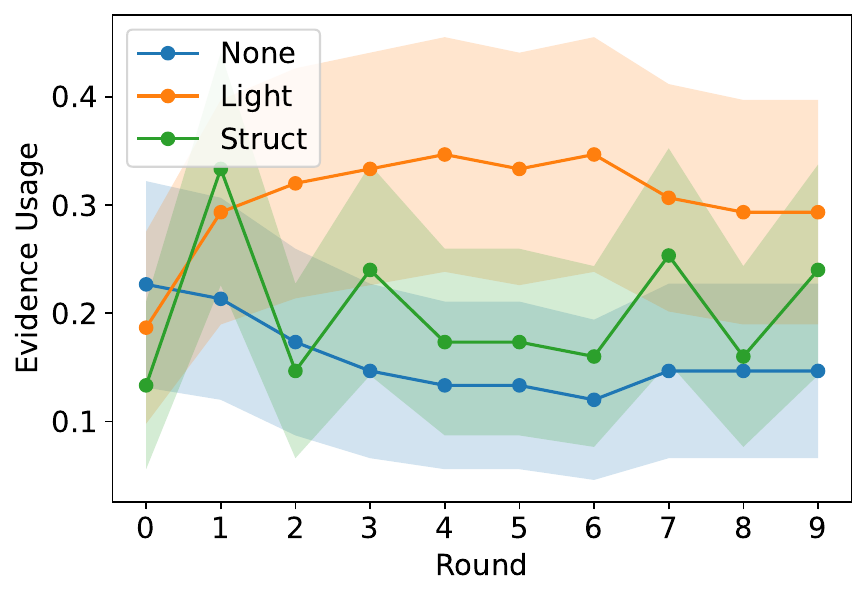}
    \caption{Evidence usage--Education}

\end{subfigure}
\hfill
\begin{subfigure}[t]{0.32\textwidth}
    \centering
    \includegraphics[width=\linewidth]{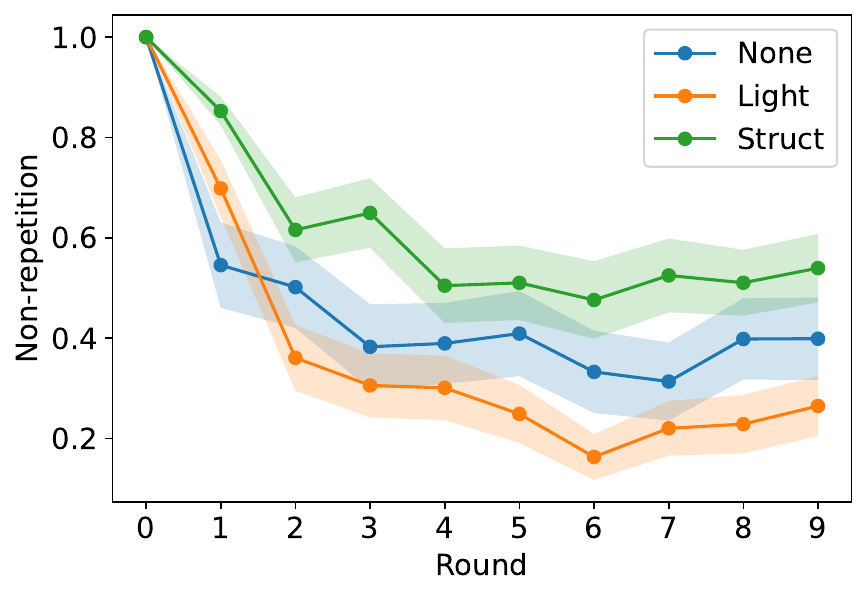}
    \caption{Non-repetition--Education}

\end{subfigure}
\caption{Round-wise changes of three dialogue metrics under different rule templates. 
Each curve shows the mean value (with 95\% confidence interval).}
\label{fig:traj_all}
\end{figure*}

\subsection{Main Study}

To discuss the research questions regarding policy parameterization effectiveness, the main study compares the performance under different \textit{Rules} and \textit{Weights} in the policy.

\subsubsection{The Effectiveness of Rules}
Based on results in Table~\ref{tab:all_metrics}, we can observe that using different rules for the same topic query can make a difference in various metrics. This shows that the agent's behaviour can be parameterised by the policy generation process (RQ1). Second, for RQ2, different rule strategies bring distinct behavioural trade-offs. For non-repetition, \textit{Struct} has the best performance, which means clearly structured rules and strong constraint prompts can reduce repetitions. For evidence usage, \textit{Light} significantly improves the performance, which suggests that \textit{Light} rules encourage the use of external knowledge, but excessive structuring may suppress evidence usage.  For rebuttal, \textit{Light} and \textit{Struct} show the higher overall rebuttal rates, reflecting more interactive and argumentative exchanges. For stance consistency, scores remain similar across all conditions (overall 0.47–0.51), implying that rule templates primarily affect interaction style rather than core issue positions. Finally, these results illustrate the potential of prompt as an interpretable influence for social simulation. 

Figure~\ref{fig:traj_all} illustrates how responsiveness, evidence usage, and non-repetition evolve across dialogue rounds. Responsiveness remains relatively stable under all three rule settings. However, for evidence usage, the degree of fluctuation increases with the level of structural constraint: the \textit{None} condition produces the smoothest trajectory, while the \textit{Struct} condition exhibits the most pronounced variation. For non-repetition, all three conditions initially decline and then converge toward a stable value, with \textit{Struct} maintaining the highest level and Light the lowest.

\subsubsection{Weights Sensitivity}

We further investigate whether different weight configurations impact performance by systematically varying each weight on the \textit{Land} scenario while holding the other two weights unchanged.
We report the overall results in the main text, with complete topic-level indicators in Appendix~\ref{app:weight}.

From Table~\ref{tab:weights}, we can observe: 1. Responsiveness remained above 0.8 regardless of weight changes, with minimal fluctuation. 2. When $W_T$ increases to 1.5, the rebuttal rate increases significantly, indicating that when persona(T) is emphasised more, agents are more likely to engage in conflict and rebuttal.  The same situation also occurs in the non-stance, high $W_T$ can also lead the agent to be more `loyal to the role' and its stance clearer and more stable. 3. About evidence usage and $W_D$, we observe a cross-over effect: when $W_D=0.5$, the \textit{Light}  condition achieves higher evidence usage, whereas when $W_D=1.5$, the None condition is comparatively stronger. This suggests that rules can enforce evidence integration even under weak weights, while in the absence of rules, stronger weights are necessary to drive evidence use.

\begin{table}[t]
\centering
\caption{Overall performance of policies with different weights and rule templates in the \textit{Land} scenario. The detailed experimental results are presented in Appendix~\ref{app:weight}.}
\label{tab:weights}
\resizebox{0.49\textwidth}{!}{
\begin{tabular}{cc|ccccc}
\toprule
\textbf{Weights} & \textbf{Rule} & \textbf{Resp.} & \textbf{Rebuttal} & \textbf{Non-rep.} & \textbf{Evid.} & \textbf{Stance} \\
\midrule
$w_T=1.0$ & None & $0.85_{\pm 0.35}$ & $0.28_{\pm 0.45}$ & $0.43_{\pm 0.42}$ & $0.10_{\pm 0.29}$ & $0.48_{\pm 0.07}$ \\
$w_M=1.0$ & Light & $0.85_{\pm 0.36}$ & $0.31_{\pm 0.46}$ & $0.33_{\pm 0.36}$ & $0.28_{\pm 0.45}$ & $0.47_{\pm 0.07}$ \\
$w_D=1.0$ & Struct & $0.80_{\pm 0.40}$ & $0.22_{\pm 0.41}$ & $0.62_{\pm 0.36}$ & $0.20_{\pm 0.40}$ & $0.47_{\pm 0.08}$ \\
\midrule
$w_T=1.0$  & None & $0.86_{\pm 0.34}$ & $0.28_{\pm 0.45}$ & $0.30_{\pm 0.35}$ & $0.30_{\pm 0.46}$ & $0.47_{\pm 0.08}$  \\
 $w_M=1.0$ & Light  & $0.84_{\pm 0.37}$ & $0.33_{\pm 0.47}$ & $0.31_{\pm 0.34}$ & $0.28_{\pm 0.45}$ & $0.44_{\pm 0.10}$ \\
 $w_D=1.5$ & Struct & $0.82_{\pm 0.38}$ & $0.17_{\pm 0.37}$ & $0.49_{\pm 0.40}$ & $0.17_{\pm 0.37}$ & $0.48_{\pm 0.07}$\\
\midrule
$w_T=1.0$ & None & $0.86_{\pm 0.35}$ & $0.28_{\pm 0.45}$ & $0.47_{\pm 0.40}$ & $0.12_{\pm 0.32}$ & $0.51_{\pm 0.08}$ \\
 $w_M=1.5$ & Light & $0.86_{\pm 0.34}$ & $0.34_{\pm 0.47}$ & $0.36_{\pm 0.36}$ & $0.26_{\pm 0.44}$ & $0.46_{\pm 0.08}$ \\
 $w_D=1.0$ & Struct & $0.82_{\pm 0.38}$ & $0.20_{\pm 0.40}$ & $0.49_{\pm 0.40}$ & $0.20_{\pm 0.40}$ & $0.49_{\pm 0.07}$ \\
\midrule
$w_T=1.5$  & None & $0.81_{\pm 0.39}$ & $0.45_{\pm 0.50}$ & $0.26_{\pm 0.37}$ & $0.05_{\pm 0.23}$ & $0.55_{\pm 0.07}$ \\
  $w_M=1.0$& Light & $0.83_{\pm 0.37}$ & $0.45_{\pm 0.50}$ & $0.30_{\pm 0.34}$ & $0.28_{\pm 0.45}$ & $0.52_{\pm 0.08}$ \\
 $w_D=1.0$ & Struct &  $0.74_{\pm 0.44}$ & $0.47_{\pm 0.50}$ & $0.36_{\pm 0.39}$ & $0.16_{\pm 0.37}$ & $0.54_{\pm 0.07}$ \\
 \midrule
$w_T=1.0$ & None &$0.87_{\pm 0.34}$ & $0.31_{\pm 0.46}$ & $0.31_{\pm 0.39}$ & $0.09_{\pm 0.29}$ & $0.48_{\pm 0.07}$ \\
 $w_M=0.5$& Light & $0.85_{\pm 0.35}$ & $0.36_{\pm 0.48}$ & $0.34_{\pm 0.35}$ & $0.32_{\pm 0.47}$ & $0.46_{\pm 0.07}$\\
 $w_D=1.0$ & Struct & $0.79_{\pm 0.41}$ & $0.23_{\pm 0.42}$ & $0.52_{\pm 0.39}$ & $0.23_{\pm 0.42}$ & $0.49_{\pm 0.07}$ \\
 \midrule
$w_T=1.0$ & None &  $0.85_{\pm 0.35}$ & $0.32_{\pm 0.47}$ & $0.41_{\pm 0.41}$ & $0.10_{\pm 0.29}$ & $0.50_{\pm 0.08}$ \\
   $w_M=1.0$& Light & $0.86_{\pm 0.35}$ & $0.31_{\pm 0.46}$ & $0.34_{\pm 0.35}$ & $0.39_{\pm 0.49}$ & $0.47_{\pm 0.08}$ \\
 $w_D=0.5$ & Struct & $0.82_{\pm 0.39}$ & $0.16_{\pm 0.36}$ & $0.57_{\pm 0.38}$ & $0.19_{\pm 0.39}$ & $0.49_{\pm 0.06}$\\
 \midrule
$w_T=0.5$ & None & $0.87_{\pm 0.34}$ & $0.32_{\pm 0.47}$ & $0.30_{\pm 0.39}$ & $0.08_{\pm 0.27}$ & $0.50_{\pm 0.07}$ \\
$w_M=1.0$ & Light & $0.88_{\pm 0.32}$ & $0.37_{\pm 0.48}$ & $0.30_{\pm 0.36}$ & $0.36_{\pm 0.48}$ & $0.47_{\pm 0.07}$ \\
  $w_D=1.0$& Struct & $0.81_{\pm 0.39}$ & $0.27_{\pm 0.44}$ & $0.49_{\pm 0.40}$ & $0.27_{\pm 0.44}$ & $0.51_{\pm 0.08}$ \\
\bottomrule

\end{tabular}
    }
\end{table}

\subsubsection{The Effectiveness of Adaptive Weights}

\begin{figure}
\begin{subfigure}[t]{0.23\textwidth}
    \centering
    \includegraphics[width=\linewidth]{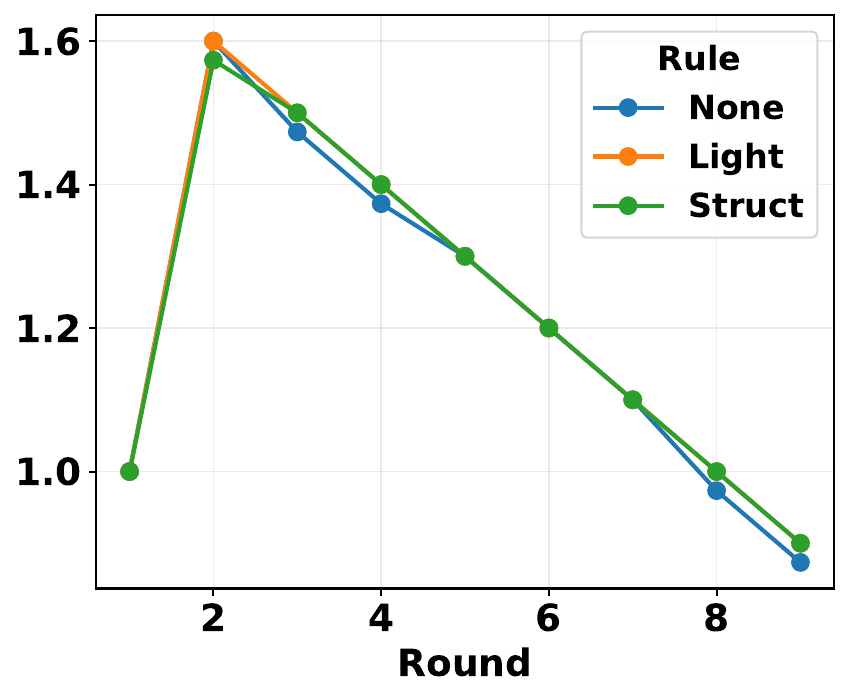}
    \caption{$W_D$}
\end{subfigure}
\begin{subfigure}[t]{0.23\textwidth}
    \centering
    \includegraphics[width=\linewidth]{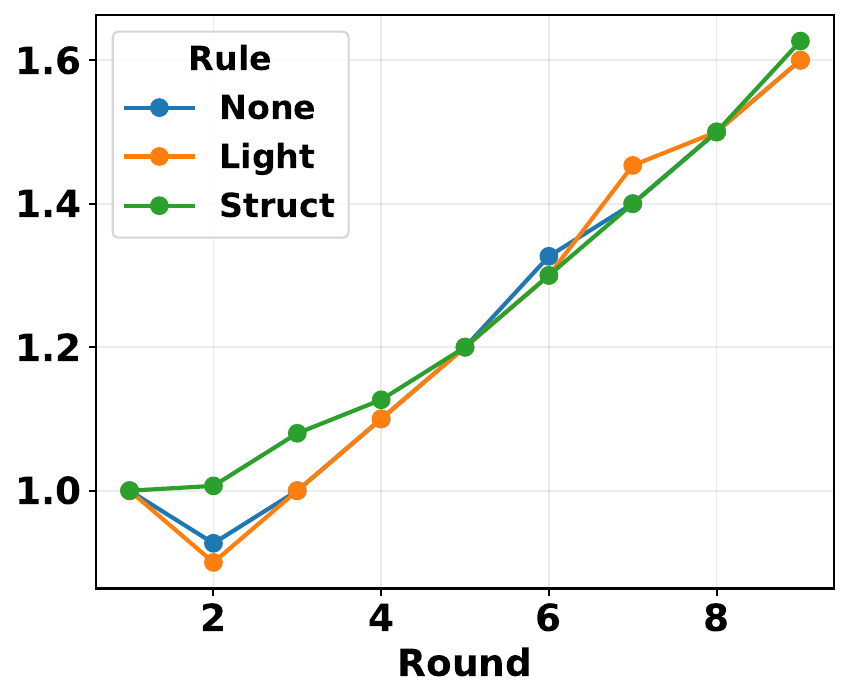}
    \caption{$W_M$}
\end{subfigure}
\caption{An Example of Adaptive Weight Changes. Farmer agent's $W_D$ and $W_M$ changes in 10 rounds under the Q1 topic.}
\label{fig:weight_change}
\end{figure}

To investigate the effectiveness of different weights in influencing agent behaviors,
we employ adaptive weights with initial parameters set as $w_T=1.0$, $w_M=1.0$, $w_D=1.0$, and $\alpha=0.2$ and conduct experiments on the \textit{Land} scenario under the same settings as the main study. We give an example of the change of $w_M$ and $w_D$ in  Figure~\ref{fig:weight_change}, and the complete experimental results are in Table~\ref{tab:adaptive}. 

\begin{figure*}[t]
\centering
\begin{subfigure}[t]{0.32\textwidth}
    \centering
    \includegraphics[width=\linewidth]{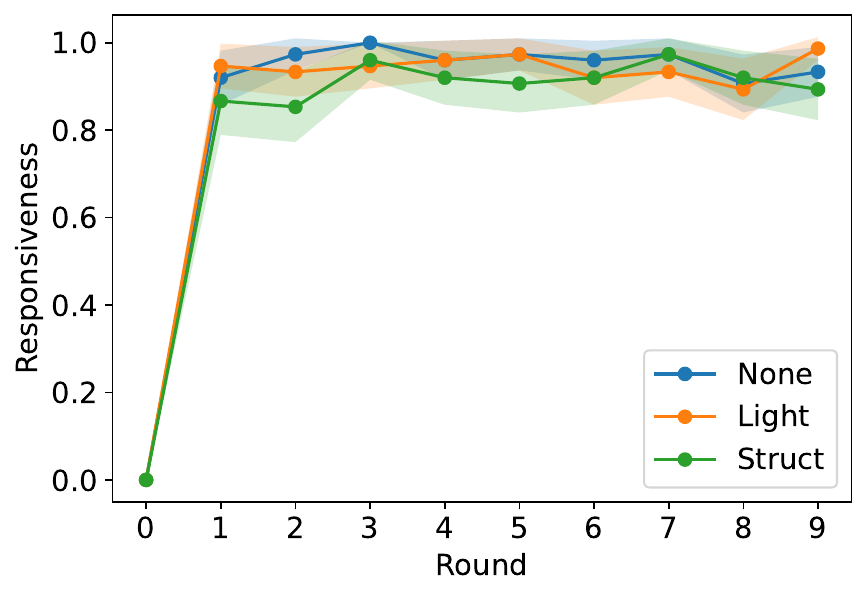}
    \caption{Responsiveness}

\end{subfigure}
\hfill
\begin{subfigure}[t]{0.32\textwidth}
    \centering
    \includegraphics[width=\linewidth]{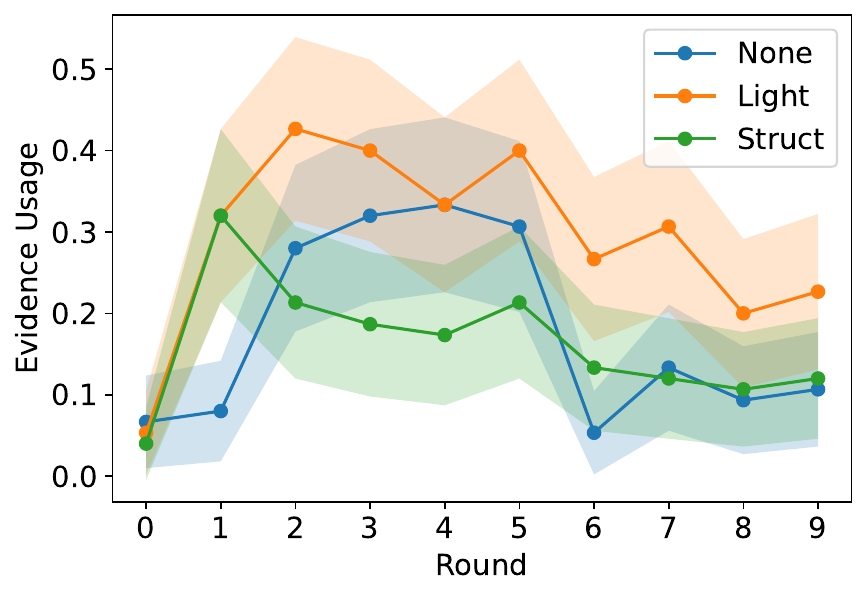}
    \caption{Evidence usage}

\end{subfigure}
\hfill
\begin{subfigure}[t]{0.32\textwidth}
    \centering
    \includegraphics[width=\linewidth]{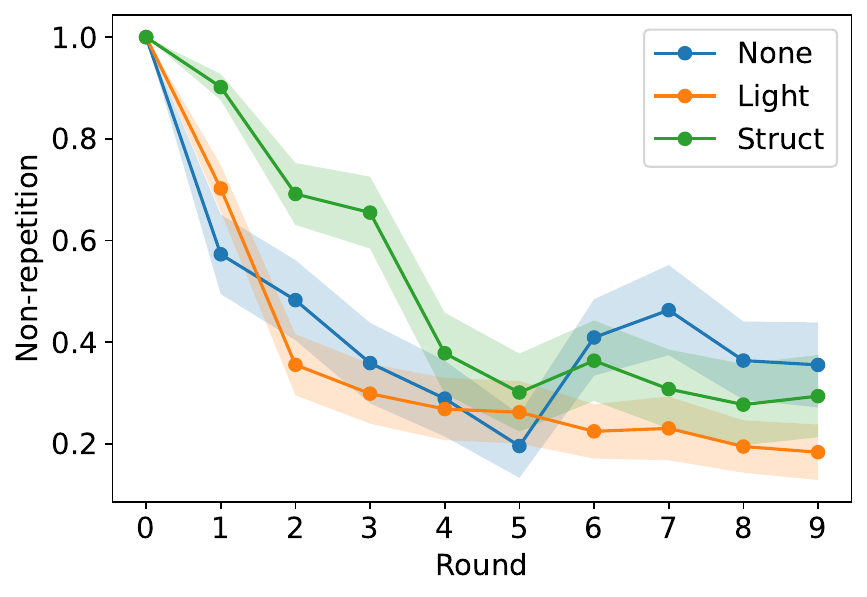}
    \caption{Non-repetition}

\end{subfigure}
\caption{Round-wise changes of three dialogue metrics with adaptive weights. }
\label{fig:adaptive_traj}
\end{figure*}

As shown in Table~\ref{tab:adaptive}, the overall averages are similar to those obtained without adaptive control, indicating that adaptive weights do not substantially alter mean performance. However, when examining round-wise trajectories (results in Figure~\ref{fig:adaptive_traj}), we observe that, for evidence usage, the curve dynamics differ from those in the main study: In the final round, the scores are lower compared to the main study, the reason is that under the adaptive setting, $w_D$ decreases over time. While in the initial rounds, the scores increase when the weight $w_D$ is set too low. This pattern is especially evident in the None condition.  These findings indicate that adaptive weights can effectively regulate the dialogue process according to their configuration.

\begin{table}[h]
\centering
\caption{Overall performance of policies with adaptive weight settings in the \textit{Land} scenario, averaged
over all agents. Detailed results are presented in Appendix~\ref{app:adaptive}.}
\label{tab:adaptive}
\resizebox{0.49\textwidth}{!}{
\begin{tabular}{cccccc}
\toprule
  \textbf{Rule} & \textbf{Resp.} & \textbf{Rebuttal} & \textbf{Non-rep.} & \textbf{Evid.} & \textbf{Stance} \\
\midrule
  None & $0.86_{\pm 0.35}$ & $0.26_{\pm 0.44}$ & $0.45_{\pm 0.39}$ & $0.18_{\pm 0.38}$ & $0.48_{\pm 0.07}$ \\
  Light & $0.85_{\pm 0.36}$ & $0.39_{\pm 0.49}$ & $0.37_{\pm 0.35}$ & $0.29_{\pm 0.46}$ & $0.45_{\pm 0.07}$ \\
  Struct & $0.82_{\pm 0.38}$ & $0.13_{\pm 0.34}$ & $0.52_{\pm 0.40}$ & $0.16_{\pm 0.37}$ & $0.48_{\pm 0.07}$ \\
\bottomrule

\end{tabular}
}
\end{table}

\subsubsection{Weight Interval and $\alpha$}
Our choice of the range for adaptive weights follows directly from our definition of the weight scale: We treat 1.0 as the mid-tier value, so weights naturally lie within $[0,2]$. One could equivalently rescale the scale to other values, but this would simply require adjusting the tier thresholds (currently 0.85 and 1.25). The thresholds and $\alpha$ operate together. Therefore, we only need to determine one parameter and adjust the other one. We chose to fix the thresholds, tested multiple $\alpha$ values in Table~\ref{tab:alpha_ablation}: the overall metrics remained stable across all settings, and $\alpha$ mainly affects the weight trajectories. Different $\alpha$ values determine how frequently weights cross tiers and how smoothly they evolve. This indicates that $\alpha$ mainly controls the smoothness of adaptation and adjusts the dialogue dynamics according to their configuration. 

\begin{table}[t]
\centering
\caption{Effect of Different $\alpha$ Values.}
\label{tab:alpha_ablation}
\resizebox{0.48\textwidth}{!}{
\begin{tabular}{c c| c c c c c}
\hline
\textbf{$\alpha$} & \textbf{Rule} & \textbf{Resp.} & \textbf{Reb.} & \textbf{Non-rep.} & \textbf{Evid.} & \textbf{Stance} \\
\toprule
0.05 & None   & $0.83\pm0.38$ & $0.24\pm0.43$ & $0.54\pm0.38$ & $0.15\pm0.35$ & $0.49\pm0.08$ \\
     & Light  & $0.83\pm0.37$ & $0.25\pm0.43$ & $0.42\pm0.36$ & $0.21\pm0.41$ & $0.46\pm0.08$ \\
     & Struct & $0.84\pm0.37$ & $0.10\pm0.30$ & $0.62\pm0.36$ & $0.18\pm0.39$ & $0.49\pm0.06$ \\
\midrule
0.1  & None   & $0.83\pm0.37$ & $0.20\pm0.40$ & $0.54\pm0.37$ & $0.17\pm0.38$ & $0.49\pm0.08$ \\
     & Light  & $0.84\pm0.37$ & $0.24\pm0.42$ & $0.41\pm0.35$ & $0.28\pm0.45$ & $0.46\pm0.08$ \\
     & Struct & $0.83\pm0.37$ & $0.08\pm0.28$ & $0.58\pm0.37$ & $0.20\pm0.40$ & $0.49\pm0.06$ \\
\midrule
0.4  & None   & $0.84\pm0.37$ & $0.24\pm0.43$ & $0.47\pm0.38$ & $0.21\pm0.41$ & $0.48\pm0.07$ \\
     & Light  & $0.84\pm0.37$ & $0.25\pm0.43$ & $0.41\pm0.36$ & $0.28\pm0.45$ & $0.46\pm0.08$ \\
     & Struct & $0.83\pm0.38$ & $0.08\pm0.27$ & $0.56\pm0.37$ & $0.20\pm0.40$ & $0.49\pm0.06$ \\
\bottomrule
\end{tabular}
}
\end{table}

\subsection{Backbone LLMs Variations}
We conduct two extra experiments on the \textit{Land} scenario with different backbone LLMs configurations.
We first test a homogeneous setup $S1$ where all agents share the same backbone LLM (Qwen3).
The results in Table~\ref{tab:backbone_vars} shows this configuration generally yields lower responsiveness, rebuttal, and non-repetition compared to our main heterogeneous setup.
This indicates that model diversity contributes to richer and more interactive discussions, whereas using a uniform backbone LLM leads to less dynamic conversational behaviour.

Second, we test an alternative heterogeneous setup to examine whether different heterogeneous backbone LLMs choices preserve the observed trends. We reassign $\langle$Llama3, Mistral, Qwen3$\rangle$ to $\langle$Farmer, Conservationist, Community Rep$\rangle$, and denote this configuration as \textit{S2}. Similarly, by reassigning $\langle$Mistral, Qwen3, Llama3$\rangle$ to the same roles, we obtain configuration \textit{S3}.
Together with the main study, these three settings (main, \textit{S2}, \textit{S3}) ensure that each agent is paired with a different backbone LLM.
The results in Table~\ref{tab:backbone_vars} show that
although the results exhibit slight variations compared to the main study, the overall conclusions remain consistent. 
Moreover, the heterogeneous backbone configuration continues to outperform the homogeneous all-Qwen3 setting, confirming that diversity in backbones LLMs leads to more robust and effective conversational behaviours.

\begin{table}[t]
\centering
\caption{Overall performance of policies comparison between homogeneous ($S1$) and heterogeneous ($S2$, $S3$) backbone LLMs variants. Results reported as mean$\pm$std. Detailed performance is in Appendix~\ref{app:backbone}.}
\label{tab:backbone_vars}
\small
\resizebox{0.48\textwidth}{!}{
\begin{tabular}{cccccccc}
\toprule
 & \textbf{Rule} & \textbf{Resp.} & \textbf{Rebuttal} & \textbf{Non-rep.} & \textbf{Evid.} & \textbf{Stance} \\
\midrule
\multirow{3}{*}{S1} & None & $0.67_{\pm 0.47}$ & $0.28_{\pm 0.45}$ & $0.25_{\pm 0.39}$ & $0.14_{\pm 0.35}$ & $0.48_{\pm 0.07}$ \\
 & Light & $0.59_{\pm 0.49}$ & $0.26_{\pm 0.44}$ & $0.26_{\pm 0.37}$ & $0.16_{\pm 0.36}$ & $0.44_{\pm 0.08}$ \\
 & Struct & $0.74_{\pm 0.44}$ & $0.12_{\pm 0.33}$ & $0.38_{\pm 0.41}$ & $0.10_{\pm 0.29}$ & $0.47_{\pm 0.07}$ \\
\midrule
\multirow{3}{*}{S2} & None & $0.71_{\pm 0.45}$ & $0.26_{\pm 0.44}$ & $0.39_{\pm 0.40}$ & $0.06_{\pm 0.23}$ & $0.48_{\pm 0.09}$ \\
 & Light & $0.72_{\pm 0.45}$ & $0.30_{\pm 0.46}$ & $0.33_{\pm 0.35}$ & $0.42_{\pm 0.49}$ & $0.46_{\pm 0.09}$ \\
 & Struct & $0.70_{\pm 0.46}$ & $0.16_{\pm 0.37}$ & $0.61_{\pm 0.36}$ & $0.22_{\pm 0.42}$ & $0.50_{\pm 0.09}$ \\
 \midrule
\multirow{3}{*}{S3} & None & $0.79_{\pm 0.41}$ & $0.23_{\pm 0.42}$ & $0.46_{\pm 0.40}$ & $0.12_{\pm 0.32}$ & $0.47_{\pm 0.08}$ \\
 & Light & $0.85_{\pm 0.36}$ & $0.24_{\pm 0.43}$ & $0.35_{\pm 0.36}$ & $0.24_{\pm 0.43}$ & $0.46_{\pm 0.07}$ \\
 & Struct & $0.84_{\pm 0.36}$ & $0.08_{\pm 0.27}$ & $0.67_{\pm 0.34}$ & $0.24_{\pm 0.43}$ & $0.50_{\pm 0.06}$ \\

 \bottomrule
\end{tabular}
}
\end{table}

In Appendix~\ref{app:more}, we provide additional experimental results on (1) the effect of different LLM backbones on agent-level behaviour, and (2) agent-level performance in larger multi-agent systems with more agents. The results show that, although agent-level performance exhibits slight variations across different backbone settings, the overall behaviours remain primarily influenced by the rules.

To avoid self-evaluation bias, the judge model is used exclusively for evaluation and is not part of the proposed framework. We further note that three out of the five evaluation metrics are embedding-based, and therefore do not rely on a judge model. However, we still evaluated the system using three independent judge models on a controlled setting of 10-round dialogues with 5 runs each. The results are highly consistent across different judges, indicating that our conclusions are robust to the choice of judge model.

\subsection{Ablation Study}

\begin{figure*}[t]
    \centering
\includegraphics[width=\linewidth]{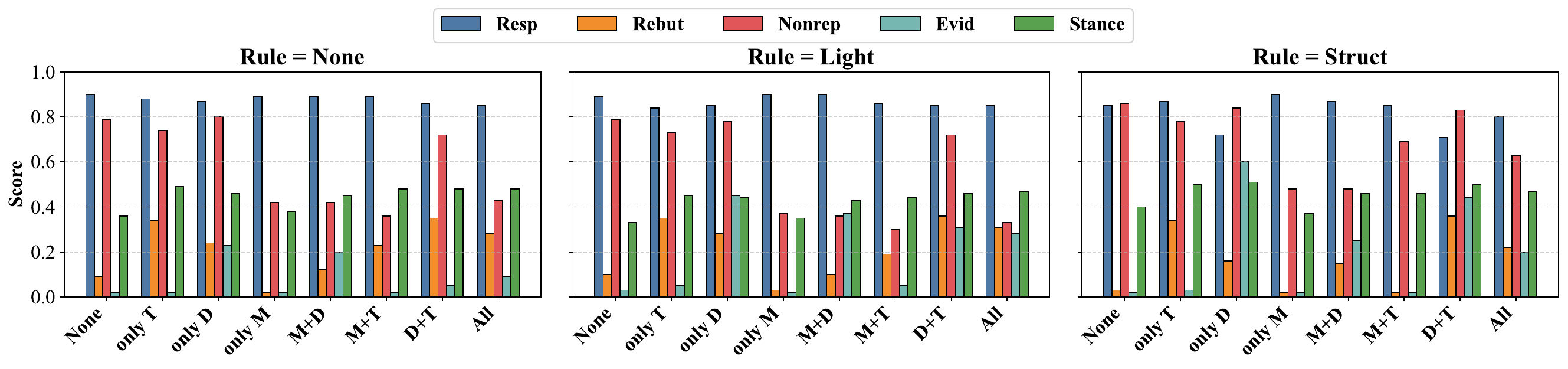}
    \caption{Ablation Study of Components T, M and D. }
    \label{fig:ablation}
\end{figure*}

Having examined the effects of the rule $R$ and weight parameters $W$ in the previous experiments, we now investigate the remaining three components, task $T$, memory $M$, and external knowledge base $D$, through a set of ablation studies conducted on the \textit{Land} scenario.
Figure~\ref{fig:ablation} reports the results of our ablation study, where we selectively remove $T$, $M$ and $D$. The results show that each component contributes in a distinct way. $T$ substantially increases rebuttal frequency and stance consistency, which verifies that a clear person and task description leads to conflict with other agents and a stable stance. $D$ increases the evidence usage, indicating that evidence retrieval can encourage grounded discussion. $M$ will cause a higher repetition, because when $M$ joins, the agent will use the information from the previous dialogue rounds to generate the response.
Combining components reveals complementarities: $D+T$ produces the most balanced performance, simultaneously improving rebuttal, evidence usage, and stance. However, enabling all three dimensions together yields only moderate improvements across metrics (like in Responsiveness, because of the use of $M$), suggesting that the effects of $T$, $M$, and $D$ can partially offset one another. Overall, these results highlight the interpretable roles of different components. We also conduct an ablation study on the EDU scenario, which supports the same conclusion as in the Land setting, and part of the experimental results are shown in Appendix~\ref{app:ablation_edu}.

\section{Discussion}

The significance of our proposed parameterized framework of the policy generation process in this study for influencing LLM-based multi-agent systems is that it redefines the role of language models in social simulations.  The language model is no longer a text generator, but rather a social actor with adjustable parameters. Through our framework, each agent's dialogue response process, which is produced by prompt-as-action, can be mapped to a set of variables with clear meanings. 
Future work could extend our framework with techniques like fine-tuning \cite{agiza2024politune} and inference-time interventions \cite{hu2025fine} to customize agent policy parameters.
This framework enables responses to be controlled by a set of clearly defined cognitive and social strategies, making social simulations more diverse and flexible, and providing a mechanism for controllable, measurable, and theoretically grounded social experiments.

\section{Conclusion}
In this study, we proposed a lightweight policy parameterised framework that regards the prompt-as-action to influence the LLM-based multi-agent dialogue. By constructing prompts through different compoents adaptively, we show that dialogue behaviours can be systematically influenced without additional training. Experiments across two scenarios demonstrated that different component settings yielded distinct effects. Overall, our findings highlight policy-parameterised prompts as an effective and interpretable mechanism for steering LLM-driven multi-agent dialogue systems, offering a promising direction for controllable social simulations.

\section{Acknowledgments}
The support of the Economic and Social Research Council (ESRC) is gratefully acknowledged. Grant Ref ES/W002639/1. Hongbo Bo is funded by ESRC Centre for Sociodigital Futures (ES/W002639/1), Jingyu Hu is funded by EPSRC-DTP ( EP/W524414/1/2894964). Weiru Liu is partially funded by ESRC Centre for Sociodigital Futures (ES/W002639/1). We thank Keri Facer for the documentation list and helpful discussion.

\clearpage

\bibliographystyle{ACM-Reference-Format} 
\bibliography{sample}

@article{lin2023agentsims,
  title={Agentsims: An open-source sandbox for large language model evaluation},
  author={Lin, Jiaju and Zhao, Haoran and Zhang, Aochi and Wu, Yiting and Ping, Huqiuyue and Chen, Qin},
  journal={arXiv preprint arXiv:2308.04026},
  year={2023}
}

@article{xu2023language,
  title={Language agents with reinforcement learning for strategic play in the werewolf game},
  author={Xu, Zelai and Yu, Chao and Fang, Fei and Wang, Yu and Wu, Yi},
  journal={arXiv preprint arXiv:2310.18940},
  year={2023}
}

@article{liu2023training,
  title={Training socially aligned language models in simulated human society},
  author={Liu, Ruibo and Yang, Ruixin and Jia, Chenyan and Zhang, Ge and Zhou, Denny and Dai, Andrew M and Yang, Diyi and Vosoughi, Soroush},
  journal={arXiv preprint arXiv:2305.16960},
  volume={2},
  year={2023}
}

@article{fan2022minedojo,
  title={Minedojo: Building open-ended embodied agents with internet-scale knowledge},
  author={Fan, Linxi and Wang, Guanzhi and Jiang, Yunfan and Mandlekar, Ajay and Yang, Yuncong and Zhu, Haoyi and Tang, Andrew and Huang, De-An and Zhu, Yuke and Anandkumar, Anima},
  journal={Advances in Neural Information Processing Systems},
  volume={35},
  pages={18343--18362},
  year={2022}
}

@inproceedings{park2023generative,
  title={Generative agents: Interactive simulacra of human behavior},
  author={Park, Joon Sung and O'Brien, Joseph and Cai, Carrie Jun and Morris, Meredith Ringel and Liang, Percy and Bernstein, Michael S},
  booktitle={Proceedings of the 36th annual acm symposium on user interface software and technology},
  pages={1--22},
  year={2023}
}

@article{gao2023s3,
  title={S3: Social-network simulation system with large language model-empowered agents},
  author={Gao, Chen and Lan, Xiaochong and Lu, Zhihong and Mao, Jinzhu and Piao, Jinghua and Wang, Huandong and Jin, Depeng and Li, Yong},
  journal={arXiv preprint arXiv:2307.14984},
  year={2023}
}

@inproceedings{park2022social,
  title={Social simulacra: Creating populated prototypes for social computing systems},
  author={Park, Joon Sung and Popowski, Lindsay and Cai, Carrie and Morris, Meredith Ringel and Liang, Percy and Bernstein, Michael S},
  booktitle={Proceedings of the 35th Annual ACM Symposium on User Interface Software and Technology},
  pages={1--18},
  year={2022}
}

@article{li2023camel,
  title={Camel: Communicative agents for" mind" exploration of large language model society},
  author={Li, Guohao and Hammoud, Hasan and Itani, Hani and Khizbullin, Dmitrii and Ghanem, Bernard},
  journal={Advances in Neural Information Processing Systems},
  volume={36},
  pages={51991--52008},
  year={2023}
}

@article{hao2023reasoning,
  title={Reasoning with language model is planning with world model},
  author={Hao, Shibo and Gu, Yi and Ma, Haodi and Hong, Joshua Jiahua and Wang, Zhen and Wang, Daisy Zhe and Hu, Zhiting},
  journal={arXiv preprint arXiv:2305.14992},
  year={2023}
}

@inproceedings{fan2024can,
  title={Can large language models serve as rational players in game theory? a systematic analysis},
  author={Fan, Caoyun and Chen, Jindou and Jin, Yaohui and He, Hao},
  booktitle={Proceedings of the AAAI Conference on Artificial Intelligence},
  volume={38},
  number={16},
  pages={17960--17967},
  year={2024}
}

@article{li2024survey,
  title={A survey on LLM-based multi-agent systems: workflow, infrastructure, and challenges},
  author={Li, Xinyi and Wang, Sai and Zeng, Siqi and Wu, Yu and Yang, Yi},
  journal={Vicinagearth},
  volume={1},
  number={1},
  pages={9},
  year={2024},
  publisher={Springer}
}

@article{guo2024large,
  title={Large language model based multi-agents: A survey of progress and challenges},
  author={Guo, Taicheng and Chen, Xiuying and Wang, Yaqi and Chang, Ruidi and Pei, Shichao and Chawla, Nitesh V and Wiest, Olaf and Zhang, Xiangliang},
  journal={arXiv preprint arXiv:2402.01680},
  year={2024}
}

@inproceedings{jang2023structured,
  title={A structured prompting based on belief-desire-intention model for proactive and explainable task planning},
  author={Jang, Minsu and Yoon, Youngwoo and Choi, Jaewoo and Ong, Hyobin and Kim, Jaehong},
  booktitle={Proceedings of the 11th International Conference on Human-Agent Interaction},
  pages={375--377},
  year={2023}
}

@article{ashery2025emergent,
  title={Emergent social conventions and collective bias in LLM populations},
  author={Ashery, Ariel Flint and Aiello, Luca Maria and Baronchelli, Andrea},
  journal={Science Advances},
  volume={11},
  number={20},
  pages={eadu9368},
  year={2025},
  publisher={American Association for the Advancement of Science}
}

@inproceedings{hong2024metagpt,
  title={MetaGPT: Meta programming for a multi-agent collaborative framework},
  author={Hong, Sirui and Zhuge, Mingchen and Chen, Jonathan and Zheng, Xiawu and Cheng, Yuheng and Zhang, Ceyao and Wang, Jinlin and Wang, Zili and Yau, Steven Ka Shing and Lin, Zijuan and others},
  year={2024},
  organization={International Conference on Learning Representations, ICLR}
}

@article{liang2023encouraging,
  title={Encouraging divergent thinking in large language models through multi-agent debate},
  author={Liang, Tian and He, Zhiwei and Jiao, Wenxiang and Wang, Xing and Wang, Yan and Wang, Rui and Yang, Yujiu and Shi, Shuming and Tu, Zhaopeng},
  journal={arXiv preprint arXiv:2305.19118},
  year={2023}
}

@article{gronauer2022multi,
  title={Multi-agent deep reinforcement learning: a survey},
  author={Gronauer, Sven and Diepold, Klaus},
  journal={Artificial Intelligence Review},
  volume={55},
  number={2},
  pages={895--943},
  year={2022},
  publisher={Springer}
}

@article{zhu2024survey,
  title={A survey of multi-agent deep reinforcement learning with communication},
  author={Zhu, Changxi and Dastani, Mehdi and Wang, Shihan},
  journal={Autonomous Agents and Multi-Agent Systems},
  volume={38},
  number={1},
  pages={4},
  year={2024},
  publisher={Springer}
}

@article{wang2020minilm,
  title={Minilm: Deep self-attention distillation for task-agnostic compression of pre-trained transformers},
  author={Wang, Wenhui and Wei, Furu and Dong, Li and Bao, Hangbo and Yang, Nan and Zhou, Ming},
  journal={Advances in neural information processing systems},
  volume={33},
  pages={5776--5788},
  year={2020}
}

@article{piao2025agentsociety,
  title={Agentsociety: Large-scale simulation of llm-driven generative agents advances understanding of human behaviors and society},
  author={Piao, Jinghua and Yan, Yuwei and Zhang, Jun and Li, Nian and Yan, Junbo and Lan, Xiaochong and Lu, Zhihong and Zheng, Zhiheng and Wang, Jing Yi and Zhou, Di and others},
  journal={arXiv preprint arXiv:2502.08691},
  year={2025}
}

@article{shinn2023reflexion,
  title={Reflexion: Language agents with verbal reinforcement learning},
  author={Shinn, Noah and Cassano, Federico and Gopinath, Ashwin and Narasimhan, Karthik and Yao, Shunyu},
  journal={Advances in Neural Information Processing Systems},
  volume={36},
  pages={8634--8652},
  year={2023}
}

@article{yang2025qwen3,
  title={Qwen3 technical report},
  author={Yang, An and Li, Anfeng and Yang, Baosong and Zhang, Beichen and Hui, Binyuan and Zheng, Bo and Yu, Bowen and Gao, Chang and Huang, Chengen and Lv, Chenxu and others},
  journal={arXiv preprint arXiv:2505.09388},
  year={2025}
}

@article{dubey2024llama,
  title={The llama 3 herd of models},
  author={Dubey, Abhimanyu and Jauhri, Abhinav and Pandey, Abhinav and Kadian, Abhishek and Al-Dahle, Ahmad and Letman, Aiesha and Mathur, Akhil and Schelten, Alan and Yang, Amy and Fan, Angela and others},
  journal={arXiv e-prints},
  pages={arXiv--2407},
  year={2024}
}

@article{jiang2023mistral7b,
  title={Mistral 7B},
  author={Jiang, Albert and von Platen, Patrick and Habib, Nazar and Le Scao, Teven and Wolf, Thomas and Mistral AI},
  journal={arXiv preprint arXiv:2310.06825},
  year={2023},
  url={https://arxiv.org/abs/2310.06825}
}

@article{lewis2020retrieval,
  title={Retrieval-augmented generation for knowledge-intensive nlp tasks},
  author={Lewis, Patrick and Perez, Ethan and Piktus, Aleksandra and Petroni, Fabio and Karpukhin, Vladimir and Goyal, Naman and K{\"u}ttler, Heinrich and Lewis, Mike and Yih, Wen-tau and Rockt{\"a}schel, Tim and others},
  journal={Advances in neural information processing systems},
  volume={33},
  pages={9459--9474},
  year={2020}
}

@article{yang2024swe,
  title={Swe-agent: Agent-computer interfaces enable automated software engineering},
  author={Yang, John and Jimenez, Carlos E and Wettig, Alexander and Lieret, Kilian and Yao, Shunyu and Narasimhan, Karthik and Press, Ofir},
  journal={Advances in Neural Information Processing Systems},
  volume={37},
  pages={50528--50652},
  year={2024}
}

@inproceedings{wu2023autogen,
  title={Autogen: Enabling next-gen LLM applications via multi-agent conversations},
  author={Wu, Qingyun and Bansal, Gagan and Zhang, Jieyu and Wu, Yiran and Li, Beibin and Zhu, Erkang and Jiang, Li and Zhang, Xiaoyun and Zhang, Shaokun and Liu, Jiale and others},
  booktitle={First Conference on Language Modeling},
  year={2024}
}

@article{estornell2024multi,
  title={Multi-LLM debate: Framework, principals, and interventions},
  author={Estornell, Andrew and Liu, Yang},
  journal={Advances in Neural Information Processing Systems},
  volume={37},
  pages={28938--28964},
  year={2024}
}

@article{abdelnabi2023llm,
  title={LLM-Deliberation: Evaluating LLMs with Interactive Multi-Agent Negotiation Games.},
  author={Abdelnabi, Sahar and Gomaa, Amr and Sivaprasad, Sarath and Sch{\"o}nherr, Lea and Fritz, Mario},
  year={2023},
  publisher={CISPA}
}

@inproceedings{agiza2024politune,
  title={Politune: Analyzing the impact of data selection and fine-tuning on economic and political biases in large language models},
  author={Agiza, Ahmed and Mostagir, Mohamed and Reda, Sherief},
  booktitle={Proceedings of the AAAI/ACM Conference on AI, Ethics, and Society},
  volume={7},
  pages={2--12},
  year={2024}
}

@article{hu2025fine,
  title={Fine-Grained Interpretation of Political Opinions in Large Language Models},
  author={Hu, Jingyu and Yang, Mengyue and Du, Mengnan and Liu, Weiru},
  journal={arXiv preprint arXiv:2506.04774},
  year={2025}
}

@inproceedings{wu2025multi,
  title={Multi-agent Deep Reinforcement Learning for Fake News Detection},
  author={Wu, Yiwen and McAreavey, Kevin and Bo, Hongbo and Liu, Weiru and McConville, Ryan},
  booktitle={2025 International Joint Conference on Neural Networks (IJCNN)},
  pages={1--8},
  year={2025},
  organization={IEEE}
}

@misc{openai2025chatgpt,
  author       = {{OpenAI}},
  title        = {ChatGPT-5},
  year         = {2025},
  howpublished = {\url{https://chat.openai.com/}},
  note         = {Large language model}
}

\clearpage

\appendix

\section{Supplement Information for Agent External Knowledge}
\label{app:materials}
The data for the \emph{Land} scenario were compiled from publicly available materials related to UK land use and farming. 
Sources include government publications such as the \emph{Land Use Futures} report
\footnote{\url{https://assets.publishing.service.gov.uk/media/5a7c31f740f0b674ed20f740/10-634-land-use-futures-summary.pdf}} 
and the recent DEFRA \emph{Land Use Framework} consultation%
\footnote{\url{https://consult.defra.gov.uk/land-use-framework/land-use-consultation/}}, 
as well as policy documents and reports from farming, conservation, and access organisations 
(e.g., the National Farmers’ Union%
\footnote{\url{https://www.nfuonline.com/updates-and-information/farming-for-britains-future-our-key-asks-of-government/}}, 
Green Alliance%
\footnote{\url{https://green-alliance.org.uk/project/the-future-of-uk-land-use/}}, 
and the Woodland Trust%
\footnote{\url{https://www.woodlandtrust.org.uk/state-of-uk-woods-and-trees/}}).

The data for the education scenario were compiled from publicly available English-language reports and international datasets on education equity, funding, and learning outcomes. Principal sources include the \emph{OECD Education at a Glance 2023}\footnote{\url{https://www.oecd.org/education/education-at-a-glance/}}, the \emph{World Bank World Development Indicators: Government Expenditure on Education (\% of GDP)}\footnote{\url{https://data.worldbank.org/indicator/SE.XPD.TOTL.GD.ZS}}, and the \emph{World Bank Open Knowledge Repository on Education Equity and Access}\footnote{\url{https://openknowledge.worldbank.org/}}. Together, these sources provide empirical evidence on spending levels, teacher distribution, rural–urban disparities, and student performance across education systems worldwide.

\section{Enhanced Instructions of Weight}
\label{app:instructions}
The detailed instructions for different weight tiers:
\begin{itemize}
    \item T
    \begin{itemize}
        \item high: 
            Speak explicitly from the [T]  perspective. First state your stance and role-specific priority, then justify with reasons.

        \item mid: 
            Reflect the [T] perspective and state your position when appropriate.
        \item low: 
            You may keep the [T]  perspective implicit; focus on arguments.
    \end{itemize}
    \item M:
    \begin{itemize}
        \item high: 
            Begin with a 1–2 sentence summary of the recent turns in [M] and address unresolved points directly.

    \item mid:  Consider the recent discussion [M] and avoid repeating earlier content.

        \item low:  You may respond without summarising [M]; avoid verbatim repetition.
    \end{itemize}
    \item D;
    \begin{itemize}
        \item high: Before concluding, list at least 2 concrete evidence points from the retrieved snippets [D]; quote/paraphrase and tie them to your claims.
        \item mid: Evidence (Preferred): Use relevant retrieved snippets [D] to support key claims when available.
        ),
        \item low
            Evidence (Optional): You may proceed without citing retrieved snippets [D] if not essential.
    \end{itemize}
\end{itemize}

\section{Weights Sensitivity}
\label{app:weight}
Tables~\ref{apptab:1},\ref{apptab:2},\ref{apptab:3},\ref{apptab:4},\ref{apptab:5} and \ref{apptab:6} are the result details of the weight sensitivity experiments.

\begin{table}[H]
\centering
\caption{$w_T=1.0, w_M=1.0, w_D=1.5$}
\label{apptab:1}
\resizebox{0.49\textwidth}{!}{
\begin{tabular}{ccccccc}
\toprule
\textbf{Query} & \textbf{Rule} & \textbf{Resp.} & \textbf{Rebuttal} & \textbf{Non-rep.} & \textbf{Evid.} & \textbf{Stance} \\
\midrule
Q1 & None & $0.90_{\pm 0.30}$ & $0.33_{\pm 0.47}$ & $0.30_{\pm 0.35}$ & $0.37_{\pm 0.48}$ & $0.52_{\pm 0.08}$ \\
 & Light & $0.83_{\pm 0.37}$ & $0.31_{\pm 0.47}$ & $0.23_{\pm 0.33}$ & $0.49_{\pm 0.50}$ & $0.47_{\pm 0.08}$ \\
 & Struct & $0.81_{\pm 0.40}$ & $0.19_{\pm 0.40}$ & $0.54_{\pm 0.40}$ & $0.29_{\pm 0.46}$ & $0.51_{\pm 0.08}$ \\
\midrule
Q2 & None & $0.90_{\pm 0.30}$ & $0.34_{\pm 0.47}$ & $0.32_{\pm 0.36}$ & $0.39_{\pm 0.49}$ & $0.46_{\pm 0.07}$ \\
 & Light & $0.88_{\pm 0.33}$ & $0.32_{\pm 0.47}$ & $0.35_{\pm 0.33}$ & $0.25_{\pm 0.44}$ & $0.40_{\pm 0.06}$ \\
 & Struct & $0.87_{\pm 0.33}$ & $0.14_{\pm 0.35}$ & $0.50_{\pm 0.40}$ & $0.17_{\pm 0.38}$ & $0.46_{\pm 0.06}$ \\
\midrule
Q3 & None & $0.90_{\pm 0.30}$ & $0.03_{\pm 0.16}$ & $0.34_{\pm 0.36}$ & $0.33_{\pm 0.47}$ & $0.47_{\pm 0.09}$ \\
 & Light & $0.89_{\pm 0.32}$ & $0.05_{\pm 0.21}$ & $0.40_{\pm 0.34}$ & $0.13_{\pm 0.34}$ & $0.42_{\pm 0.12}$ \\
 & Struct & $0.89_{\pm 0.31}$ & $0.02_{\pm 0.14}$ & $0.52_{\pm 0.38}$ & $0.11_{\pm 0.32}$ & $0.45_{\pm 0.07}$ \\
\midrule
Q4 & None & $0.73_{\pm 0.45}$ & $0.53_{\pm 0.50}$ & $0.29_{\pm 0.35}$ & $0.15_{\pm 0.35}$ & $0.50_{\pm 0.08}$ \\
 & Light & $0.71_{\pm 0.46}$ & $0.51_{\pm 0.50}$ & $0.28_{\pm 0.34}$ & $0.23_{\pm 0.42}$ & $0.49_{\pm 0.10}$ \\
 & Struct & $0.69_{\pm 0.46}$ & $0.33_{\pm 0.47}$ & $0.40_{\pm 0.40}$ & $0.12_{\pm 0.33}$ & $0.49_{\pm 0.06}$ \\
\midrule
Q5 & None & $0.89_{\pm 0.32}$ & $0.17_{\pm 0.37}$ & $0.27_{\pm 0.34}$ & $0.24_{\pm 0.43}$ & $0.42_{\pm 0.04}$ \\
 & Light & $0.89_{\pm 0.32}$ & $0.45_{\pm 0.50}$ & $0.29_{\pm 0.34}$ & $0.28_{\pm 0.45}$ & $0.41_{\pm 0.05}$ \\
 & Struct & $0.86_{\pm 0.35}$ & $0.15_{\pm 0.35}$ & $0.48_{\pm 0.39}$ & $0.13_{\pm 0.33}$ & $0.46_{\pm 0.07}$ \\
\midrule
Overall & None & $0.86_{\pm 0.34}$ & $0.28_{\pm 0.45}$ & $0.30_{\pm 0.35}$ & $0.30_{\pm 0.46}$ & $0.47_{\pm 0.08}$ \\
 & Light & $0.84_{\pm 0.37}$ & $0.33_{\pm 0.47}$ & $0.31_{\pm 0.34}$ & $0.28_{\pm 0.45}$ & $0.44_{\pm 0.10}$ \\
 & Struct & $0.82_{\pm 0.38}$ & $0.17_{\pm 0.37}$ & $0.49_{\pm 0.40}$ & $0.17_{\pm 0.37}$ & $0.48_{\pm 0.07}$ \\
\bottomrule

\end{tabular}
}
\end{table}

\begin{table}[H]
\centering
\caption{$w_T=1.0, w_M=1.5, w_D=1.0$}
\label{apptab:2}
\resizebox{0.49\textwidth}{!}{
\begin{tabular}{ccccccc}
\toprule
\textbf{Query} & \textbf{Rule} & \textbf{Resp.} & \textbf{Rebuttal} & \textbf{Non-rep.} & \textbf{Evid.} & \textbf{Stance} \\
\midrule
Q1 & None & $0.86_{\pm 0.35}$ & $0.27_{\pm 0.45}$ & $0.57_{\pm 0.37}$ & $0.27_{\pm 0.44}$ & $0.56_{\pm 0.08}$ \\
 & Light & $0.88_{\pm 0.33}$ & $0.44_{\pm 0.50}$ & $0.38_{\pm 0.37}$ & $0.38_{\pm 0.49}$ & $0.49_{\pm 0.07}$ \\
 & Struct & $0.83_{\pm 0.37}$ & $0.22_{\pm 0.42}$ & $0.53_{\pm 0.40}$ & $0.35_{\pm 0.48}$ & $0.51_{\pm 0.08}$ \\
\midrule
Q2 & None & $0.87_{\pm 0.33}$ & $0.37_{\pm 0.48}$ & $0.46_{\pm 0.41}$ & $0.05_{\pm 0.23}$ & $0.49_{\pm 0.06}$ \\
 & Light & $0.88_{\pm 0.33}$ & $0.28_{\pm 0.45}$ & $0.36_{\pm 0.37}$ & $0.23_{\pm 0.42}$ & $0.43_{\pm 0.06}$ \\
 & Struct & $0.86_{\pm 0.35}$ & $0.25_{\pm 0.43}$ & $0.47_{\pm 0.41}$ & $0.12_{\pm 0.33}$ & $0.49_{\pm 0.06}$ \\
\midrule
Q3 & None & $0.89_{\pm 0.31}$ & $0.01_{\pm 0.12}$ & $0.52_{\pm 0.38}$ & $0.09_{\pm 0.29}$ & $0.49_{\pm 0.06}$ \\
 & Light & $0.87_{\pm 0.33}$ & $0.09_{\pm 0.29}$ & $0.45_{\pm 0.36}$ & $0.09_{\pm 0.29}$ & $0.43_{\pm 0.06}$ \\
 & Struct & $0.88_{\pm 0.33}$ & $0.01_{\pm 0.08}$ & $0.50_{\pm 0.40}$ & $0.10_{\pm 0.30}$ & $0.47_{\pm 0.06}$ \\
\midrule
Q4 & None & $0.76_{\pm 0.43}$ & $0.45_{\pm 0.50}$ & $0.30_{\pm 0.38}$ & $0.06_{\pm 0.24}$ & $0.51_{\pm 0.08}$ \\
 & Light & $0.81_{\pm 0.39}$ & $0.56_{\pm 0.50}$ & $0.28_{\pm 0.34}$ & $0.39_{\pm 0.49}$ & $0.50_{\pm 0.09}$ \\
 & Struct & $0.70_{\pm 0.46}$ & $0.37_{\pm 0.48}$ & $0.41_{\pm 0.40}$ & $0.29_{\pm 0.45}$ & $0.52_{\pm 0.06}$ \\
\midrule
Q5 & None & $0.89_{\pm 0.31}$ & $0.29_{\pm 0.45}$ & $0.48_{\pm 0.40}$ & $0.12_{\pm 0.33}$ & $0.48_{\pm 0.09}$ \\
 & Light & $0.87_{\pm 0.34}$ & $0.33_{\pm 0.47}$ & $0.36_{\pm 0.34}$ & $0.17_{\pm 0.38}$ & $0.42_{\pm 0.07}$ \\
 & Struct & $0.85_{\pm 0.36}$ & $0.13_{\pm 0.33}$ & $0.56_{\pm 0.36}$ & $0.15_{\pm 0.35}$ & $0.47_{\pm 0.06}$ \\
\midrule
Overall & None & $0.86_{\pm 0.35}$ & $0.28_{\pm 0.45}$ & $0.47_{\pm 0.40}$ & $0.12_{\pm 0.32}$ & $0.51_{\pm 0.08}$ \\
 & Light & $0.86_{\pm 0.34}$ & $0.34_{\pm 0.47}$ & $0.36_{\pm 0.36}$ & $0.26_{\pm 0.44}$ & $0.46_{\pm 0.08}$ \\
 & Struct & $0.82_{\pm 0.38}$ & $0.20_{\pm 0.40}$ & $0.49_{\pm 0.40}$ & $0.20_{\pm 0.40}$ & $0.49_{\pm 0.07}$ \\
\bottomrule

\end{tabular}
}
\end{table}

\begin{table}[H]
\centering
\caption{$w_T=1.5, w_M=1.0, w_D=1.0$}
\label{apptab:3}
\resizebox{0.49\textwidth}{!}{
\begin{tabular}{ccccccc}
\toprule
\textbf{Query} & \textbf{Rule} & \textbf{Resp.} & \textbf{Rebuttal} & \textbf{Non-rep.} & \textbf{Evid.} & \textbf{Stance} \\
\midrule
Q1 & None & $0.88_{\pm 0.33}$ & $0.33_{\pm 0.47}$ & $0.32_{\pm 0.39}$ & $0.08_{\pm 0.27}$ & $0.61_{\pm 0.05}$ \\
 & Light & $0.83_{\pm 0.38}$ & $0.41_{\pm 0.49}$ & $0.27_{\pm 0.33}$ & $0.34_{\pm 0.47}$ & $0.56_{\pm 0.07}$ \\
 & Struct & $0.90_{\pm 0.30}$ & $0.27_{\pm 0.45}$ & $0.37_{\pm 0.38}$ & $0.25_{\pm 0.44}$ & $0.59_{\pm 0.07}$ \\
\midrule
Q2 & None & $0.86_{\pm 0.35}$ & $0.30_{\pm 0.46}$ & $0.21_{\pm 0.34}$ & $0.00_{\pm 0.00}$ & $0.54_{\pm 0.06}$ \\
 & Light & $0.89_{\pm 0.31}$ & $0.31_{\pm 0.46}$ & $0.26_{\pm 0.33}$ & $0.25_{\pm 0.44}$ & $0.50_{\pm 0.06}$ \\
 & Struct & $0.79_{\pm 0.41}$ & $0.48_{\pm 0.50}$ & $0.29_{\pm 0.37}$ & $0.08_{\pm 0.27}$ & $0.51_{\pm 0.05}$ \\
\midrule
Q3 & None & $0.72_{\pm 0.45}$ & $0.41_{\pm 0.49}$ & $0.23_{\pm 0.37}$ & $0.01_{\pm 0.08}$ & $0.55_{\pm 0.07}$ \\
 & Light & $0.84_{\pm 0.37}$ & $0.31_{\pm 0.47}$ & $0.36_{\pm 0.36}$ & $0.29_{\pm 0.45}$ & $0.53_{\pm 0.09}$ \\
 & Struct & $0.63_{\pm 0.48}$ & $0.45_{\pm 0.50}$ & $0.40_{\pm 0.41}$ & $0.14_{\pm 0.35}$ & $0.52_{\pm 0.07}$ \\
\midrule
Q4 & None & $0.75_{\pm 0.44}$ & $0.60_{\pm 0.49}$ & $0.21_{\pm 0.35}$ & $0.15_{\pm 0.35}$ & $0.55_{\pm 0.07}$ \\
 & Light & $0.75_{\pm 0.44}$ & $0.60_{\pm 0.49}$ & $0.26_{\pm 0.35}$ & $0.31_{\pm 0.46}$ & $0.53_{\pm 0.06}$ \\
 & Struct & $0.61_{\pm 0.49}$ & $0.57_{\pm 0.50}$ & $0.33_{\pm 0.38}$ & $0.15_{\pm 0.36}$ & $0.54_{\pm 0.07}$ \\
\midrule
Q5 & None & $0.86_{\pm 0.35}$ & $0.61_{\pm 0.49}$ & $0.32_{\pm 0.37}$ & $0.03_{\pm 0.18}$ & $0.51_{\pm 0.07}$ \\
 & Light & $0.85_{\pm 0.35}$ & $0.61_{\pm 0.49}$ & $0.35_{\pm 0.35}$ & $0.19_{\pm 0.40}$ & $0.47_{\pm 0.07}$ \\
 & Struct & $0.79_{\pm 0.41}$ & $0.56_{\pm 0.50}$ & $0.39_{\pm 0.38}$ & $0.17_{\pm 0.38}$ & $0.53_{\pm 0.06}$ \\
\midrule
Overall & None & $0.81_{\pm 0.39}$ & $0.45_{\pm 0.50}$ & $0.26_{\pm 0.37}$ & $0.05_{\pm 0.23}$ & $0.55_{\pm 0.07}$ \\
 & Light & $0.83_{\pm 0.37}$ & $0.45_{\pm 0.50}$ & $0.30_{\pm 0.34}$ & $0.28_{\pm 0.45}$ & $0.52_{\pm 0.08}$ \\
 & Struct & $0.74_{\pm 0.44}$ & $0.47_{\pm 0.50}$ & $0.36_{\pm 0.39}$ & $0.16_{\pm 0.37}$ & $0.54_{\pm 0.07}$ \\
\bottomrule

\end{tabular}
}
\end{table}

\begin{table}[H]
\centering
\caption{$w_T=1.0, w_M=0.5, w_D=1.0$}
\label{apptab:4}
\resizebox{0.49\textwidth}{!}{
\begin{tabular}{ccccccc}
\toprule
\textbf{Query} & \textbf{Rule} & \textbf{Resp.} & \textbf{Rebuttal} & \textbf{Non-rep.} & \textbf{Evid.} & \textbf{Stance} \\
\midrule
Q1 & None & $0.88_{\pm 0.33}$ & $0.31_{\pm 0.46}$ & $0.37_{\pm 0.41}$ & $0.23_{\pm 0.42}$ & $0.56_{\pm 0.05}$ \\
 & Light & $0.87_{\pm 0.34}$ & $0.35_{\pm 0.48}$ & $0.40_{\pm 0.37}$ & $0.33_{\pm 0.47}$ & $0.51_{\pm 0.08}$ \\
 & Struct & $0.75_{\pm 0.43}$ & $0.26_{\pm 0.44}$ & $0.53_{\pm 0.40}$ & $0.35_{\pm 0.48}$ & $0.51_{\pm 0.08}$ \\
\midrule
Q2 & None & $0.89_{\pm 0.31}$ & $0.44_{\pm 0.50}$ & $0.29_{\pm 0.39}$ & $0.08_{\pm 0.27}$ & $0.46_{\pm 0.05}$ \\
 & Light & $0.90_{\pm 0.30}$ & $0.31_{\pm 0.46}$ & $0.35_{\pm 0.34}$ & $0.21_{\pm 0.41}$ & $0.44_{\pm 0.04}$ \\
 & Struct & $0.87_{\pm 0.34}$ & $0.21_{\pm 0.41}$ & $0.52_{\pm 0.39}$ & $0.31_{\pm 0.47}$ & $0.51_{\pm 0.07}$ \\
\midrule
Q3 & None & $0.90_{\pm 0.30}$ & $0.00_{\pm 0.00}$ & $0.38_{\pm 0.42}$ & $0.05_{\pm 0.23}$ & $0.46_{\pm 0.06}$ \\
 & Light & $0.90_{\pm 0.30}$ & $0.03_{\pm 0.16}$ & $0.36_{\pm 0.37}$ & $0.23_{\pm 0.42}$ & $0.45_{\pm 0.06}$ \\
 & Struct & $0.89_{\pm 0.32}$ & $0.01_{\pm 0.12}$ & $0.60_{\pm 0.36}$ & $0.09_{\pm 0.29}$ & $0.46_{\pm 0.07}$ \\
\midrule
Q4 & None & $0.78_{\pm 0.42}$ & $0.55_{\pm 0.50}$ & $0.29_{\pm 0.39}$ & $0.07_{\pm 0.26}$ & $0.50_{\pm 0.05}$ \\
 & Light & $0.70_{\pm 0.46}$ & $0.57_{\pm 0.50}$ & $0.25_{\pm 0.35}$ & $0.67_{\pm 0.47}$ & $0.49_{\pm 0.09}$ \\
 & Struct & $0.67_{\pm 0.47}$ & $0.51_{\pm 0.50}$ & $0.39_{\pm 0.40}$ & $0.22_{\pm 0.42}$ & $0.51_{\pm 0.05}$ \\
\midrule
Q5 & None & $0.89_{\pm 0.31}$ & $0.25_{\pm 0.44}$ & $0.24_{\pm 0.35}$ & $0.03_{\pm 0.16}$ & $0.44_{\pm 0.03}$ \\
 & Light & $0.89_{\pm 0.31}$ & $0.54_{\pm 0.50}$ & $0.32_{\pm 0.34}$ & $0.15_{\pm 0.35}$ & $0.42_{\pm 0.04}$ \\
 & Struct & $0.77_{\pm 0.42}$ & $0.18_{\pm 0.39}$ & $0.54_{\pm 0.40}$ & $0.17_{\pm 0.37}$ & $0.46_{\pm 0.07}$ \\
\midrule
Overall & None & $0.87_{\pm 0.34}$ & $0.31_{\pm 0.46}$ & $0.31_{\pm 0.39}$ & $0.09_{\pm 0.29}$ & $0.48_{\pm 0.07}$ \\
 & Light & $0.85_{\pm 0.35}$ & $0.36_{\pm 0.48}$ & $0.34_{\pm 0.35}$ & $0.32_{\pm 0.47}$ & $0.46_{\pm 0.07}$ \\
 & Struct & $0.79_{\pm 0.41}$ & $0.23_{\pm 0.42}$ & $0.52_{\pm 0.39}$ & $0.23_{\pm 0.42}$ & $0.49_{\pm 0.07}$ \\
\bottomrule

\end{tabular}
}
\end{table}

\begin{table}[H]
\centering
\caption{$w_T=1.0, w_M=1.0, w_D=0.5$}
\label{apptab:5}
\resizebox{0.49\textwidth}{!}{
\begin{tabular}{ccccccc}
\toprule
\textbf{Query} & \textbf{Rule} & \textbf{Resp.} & \textbf{Rebuttal} & \textbf{Non-rep.} & \textbf{Evid.} & \textbf{Stance} \\
\midrule
Q1 & None & $0.86_{\pm 0.35}$ & $0.19_{\pm 0.39}$ & $0.42_{\pm 0.41}$ & $0.20_{\pm 0.40}$ & $0.54_{\pm 0.08}$ \\
 & Light & $0.88_{\pm 0.33}$ & $0.31_{\pm 0.47}$ & $0.33_{\pm 0.35}$ & $0.55_{\pm 0.50}$ & $0.51_{\pm 0.10}$ \\
 & Struct & $0.76_{\pm 0.43}$ & $0.31_{\pm 0.46}$ & $0.57_{\pm 0.38}$ & $0.42_{\pm 0.49}$ & $0.52_{\pm 0.07}$ \\
\midrule
Q2 & None & $0.82_{\pm 0.39}$ & $0.50_{\pm 0.50}$ & $0.45_{\pm 0.42}$ & $0.11_{\pm 0.32}$ & $0.49_{\pm 0.08}$ \\
 & Light & $0.87_{\pm 0.34}$ & $0.33_{\pm 0.47}$ & $0.32_{\pm 0.34}$ & $0.18_{\pm 0.39}$ & $0.44_{\pm 0.05}$ \\
 & Struct & $0.79_{\pm 0.41}$ & $0.12_{\pm 0.33}$ & $0.57_{\pm 0.37}$ & $0.05_{\pm 0.23}$ & $0.48_{\pm 0.06}$ \\
\midrule
Q3 & None & $0.89_{\pm 0.32}$ & $0.02_{\pm 0.14}$ & $0.44_{\pm 0.42}$ & $0.06_{\pm 0.24}$ & $0.51_{\pm 0.10}$ \\
 & Light & $0.89_{\pm 0.32}$ & $0.01_{\pm 0.12}$ & $0.41_{\pm 0.37}$ & $0.15_{\pm 0.36}$ & $0.48_{\pm 0.09}$ \\
 & Struct & $0.85_{\pm 0.36}$ & $0.01_{\pm 0.12}$ & $0.60_{\pm 0.36}$ & $0.19_{\pm 0.40}$ & $0.47_{\pm 0.06}$ \\
\midrule
Q4 & None & $0.81_{\pm 0.39}$ & $0.54_{\pm 0.50}$ & $0.28_{\pm 0.36}$ & $0.04_{\pm 0.20}$ & $0.49_{\pm 0.05}$ \\
 & Light & $0.81_{\pm 0.39}$ & $0.45_{\pm 0.50}$ & $0.28_{\pm 0.35}$ & $0.64_{\pm 0.48}$ & $0.50_{\pm 0.07}$ \\
 & Struct & $0.87_{\pm 0.34}$ & $0.22_{\pm 0.42}$ & $0.54_{\pm 0.38}$ & $0.12_{\pm 0.33}$ & $0.49_{\pm 0.05}$ \\
\midrule
Q5 & None & $0.89_{\pm 0.31}$ & $0.37_{\pm 0.48}$ & $0.44_{\pm 0.39}$ & $0.06_{\pm 0.24}$ & $0.47_{\pm 0.07}$ \\
 & Light & $0.85_{\pm 0.35}$ & $0.42_{\pm 0.49}$ & $0.33_{\pm 0.35}$ & $0.45_{\pm 0.50}$ & $0.44_{\pm 0.06}$ \\
 & Struct & $0.83_{\pm 0.38}$ & $0.12_{\pm 0.33}$ & $0.56_{\pm 0.39}$ & $0.17_{\pm 0.38}$ & $0.49_{\pm 0.07}$ \\
\midrule
Overall & None & $0.85_{\pm 0.35}$ & $0.32_{\pm 0.47}$ & $0.41_{\pm 0.41}$ & $0.10_{\pm 0.29}$ & $0.50_{\pm 0.08}$ \\
 & Light & $0.86_{\pm 0.35}$ & $0.31_{\pm 0.46}$ & $0.34_{\pm 0.35}$ & $0.39_{\pm 0.49}$ & $0.47_{\pm 0.08}$ \\
 & Struct & $0.82_{\pm 0.39}$ & $0.16_{\pm 0.36}$ & $0.57_{\pm 0.38}$ & $0.19_{\pm 0.39}$ & $0.49_{\pm 0.06}$ \\
\bottomrule

\end{tabular}
}
\end{table}

\begin{table}[H]
\centering
\caption{$w_T=0.5, w_M=1.0, w_D=1.0$}
\label{apptab:6}
\resizebox{0.49\textwidth}{!}{
\begin{tabular}{ccccccc}
\toprule
\textbf{Query} & \textbf{Rule} & \textbf{Resp.} & \textbf{Rebuttal} & \textbf{Non-rep.} & \textbf{Evid.} & \textbf{Stance} \\
\midrule
Q1 & None & $0.89_{\pm 0.31}$ & $0.31_{\pm 0.46}$ & $0.36_{\pm 0.41}$ & $0.18_{\pm 0.39}$ & $0.56_{\pm 0.07}$ \\
 & Light & $0.90_{\pm 0.30}$ & $0.51_{\pm 0.50}$ & $0.32_{\pm 0.36}$ & $0.27_{\pm 0.44}$ & $0.49_{\pm 0.08}$ \\
 & Struct & $0.83_{\pm 0.37}$ & $0.33_{\pm 0.47}$ & $0.55_{\pm 0.38}$ & $0.34_{\pm 0.47}$ & $0.57_{\pm 0.08}$ \\
\midrule
Q2 & None & $0.89_{\pm 0.32}$ & $0.38_{\pm 0.49}$ & $0.29_{\pm 0.39}$ & $0.03_{\pm 0.16}$ & $0.48_{\pm 0.06}$ \\
 & Light & $0.90_{\pm 0.30}$ & $0.30_{\pm 0.46}$ & $0.22_{\pm 0.33}$ & $0.46_{\pm 0.50}$ & $0.43_{\pm 0.05}$ \\
 & Struct & $0.85_{\pm 0.36}$ & $0.20_{\pm 0.40}$ & $0.51_{\pm 0.40}$ & $0.12_{\pm 0.33}$ & $0.49_{\pm 0.06}$ \\
\midrule
Q3 & None & $0.87_{\pm 0.33}$ & $0.01_{\pm 0.08}$ & $0.31_{\pm 0.40}$ & $0.04_{\pm 0.20}$ & $0.50_{\pm 0.07}$ \\
 & Light & $0.90_{\pm 0.30}$ & $0.02_{\pm 0.14}$ & $0.39_{\pm 0.39}$ & $0.32_{\pm 0.47}$ & $0.48_{\pm 0.09}$ \\
 & Struct & $0.87_{\pm 0.33}$ & $0.06_{\pm 0.24}$ & $0.54_{\pm 0.40}$ & $0.25_{\pm 0.44}$ & $0.49_{\pm 0.08}$ \\
\midrule
Q4 & None & $0.79_{\pm 0.41}$ & $0.60_{\pm 0.49}$ & $0.24_{\pm 0.37}$ & $0.13_{\pm 0.34}$ & $0.51_{\pm 0.06}$ \\
 & Light & $0.83_{\pm 0.37}$ & $0.53_{\pm 0.50}$ & $0.29_{\pm 0.36}$ & $0.58_{\pm 0.49}$ & $0.49_{\pm 0.07}$ \\
 & Struct & $0.71_{\pm 0.45}$ & $0.44_{\pm 0.50}$ & $0.39_{\pm 0.39}$ & $0.50_{\pm 0.50}$ & $0.51_{\pm 0.07}$ \\
\midrule
Q5 & None & $0.89_{\pm 0.31}$ & $0.32_{\pm 0.47}$ & $0.29_{\pm 0.37}$ & $0.02_{\pm 0.14}$ & $0.42_{\pm 0.03}$ \\
 & Light & $0.88_{\pm 0.33}$ & $0.51_{\pm 0.50}$ & $0.27_{\pm 0.34}$ & $0.19_{\pm 0.40}$ & $0.42_{\pm 0.04}$ \\
 & Struct & $0.80_{\pm 0.40}$ & $0.32_{\pm 0.47}$ & $0.49_{\pm 0.41}$ & $0.13_{\pm 0.34}$ & $0.49_{\pm 0.07}$ \\
\midrule
Overall & None & $0.87_{\pm 0.34}$ & $0.32_{\pm 0.47}$ & $0.30_{\pm 0.39}$ & $0.08_{\pm 0.27}$ & $0.50_{\pm 0.07}$ \\
 & Light & $0.88_{\pm 0.32}$ & $0.37_{\pm 0.48}$ & $0.30_{\pm 0.36}$ & $0.36_{\pm 0.48}$ & $0.47_{\pm 0.07}$ \\
 & Struct & $0.81_{\pm 0.39}$ & $0.27_{\pm 0.44}$ & $0.49_{\pm 0.40}$ & $0.27_{\pm 0.44}$ & $0.51_{\pm 0.08}$ \\
\bottomrule

\end{tabular}
}
\end{table}

\section{Ablation Study}
Table~\ref{apptab:7} provides the details of the ablation study.

\begin{table}[H]
\centering
\caption{Ablation Study}
\label{apptab:7}
\resizebox{0.49\textwidth}{!}{
\begin{tabular}{cc|ccccc}
\toprule
\textbf{} & \textbf{Rule} & \textbf{Resp.} & \textbf{Rebuttal} & \textbf{Non-rep.} & \textbf{Evid.} & \textbf{Stance} \\
\midrule
None & None & $0.90_{\pm 0.30}$ & $0.09_{\pm 0.29}$ & $0.79_{\pm 0.16}$ & $0.00_{\pm 0.00}$ & $0.36_{\pm 0.09}$ \\
 & Light & $0.89_{\pm 0.31}$ & $0.10_{\pm 0.30}$ & $0.79_{\pm 0.14}$ & $0.03_{\pm 0.16}$ & $0.33_{\pm 0.08}$ \\
 & Struct & $0.85_{\pm 0.35}$ & $0.03_{\pm 0.17}$ & $0.86_{\pm 0.11}$ & $0.01_{\pm 0.08}$ & $0.40_{\pm 0.08}$ \\
 \midrule
only T& None & $0.88_{\pm 0.33}$ & $0.34_{\pm 0.47}$ & $0.74_{\pm 0.20}$ & $0.00_{\pm 0.05}$ & $0.49_{\pm 0.09}$ \\
 & Light & $0.84_{\pm 0.37}$ & $0.35_{\pm 0.48}$ & $0.73_{\pm 0.23}$ & $0.05_{\pm 0.21}$ & $0.45_{\pm 0.09}$ \\
 & Struct & $0.87_{\pm 0.33}$ & $0.34_{\pm 0.47}$ & $0.78_{\pm 0.18}$ & $0.03_{\pm 0.17}$ & $0.50_{\pm 0.08}$ \\
\midrule
only D & None & $0.87_{\pm 0.34}$ & $0.24_{\pm 0.43}$ & $0.80_{\pm 0.15}$ & $0.23_{\pm 0.42}$ & $0.46_{\pm 0.08}$ \\
 & Light & $0.85_{\pm 0.35}$ & $0.28_{\pm 0.45}$ & $0.78_{\pm 0.14}$ & $0.45_{\pm 0.50}$ & $0.44_{\pm 0.08}$ \\
 & Struct & $0.72_{\pm 0.45}$ & $0.16_{\pm 0.36}$ & $0.84_{\pm 0.14}$ & $0.60_{\pm 0.49}$ & $0.51_{\pm 0.06}$ \\
\midrule
only M & None & $0.89_{\pm 0.31}$ & $0.00_{\pm 0.06}$ & $0.42_{\pm 0.37}$ & $0.00_{\pm 0.06}$ & $0.38_{\pm 0.08}$ \\
 & Light & $0.90_{\pm 0.30}$ & $0.03_{\pm 0.17}$ & $0.37_{\pm 0.36}$ & $0.01_{\pm 0.12}$ & $0.35_{\pm 0.08}$ \\
 & Struct & $0.90_{\pm 0.30}$ & $0.01_{\pm 0.10}$ & $0.48_{\pm 0.38}$ & $0.01_{\pm 0.08}$ & $0.37_{\pm 0.09}$ \\
 \midrule
M+D &  None & $0.90_{\pm 0.31}$ & $0.12_{\pm 0.33}$ & $0.42_{\pm 0.40}$ & $0.20_{\pm 0.40}$ & $0.45_{\pm 0.09}$ \\
 & Light & $0.90_{\pm 0.30}$ & $0.10_{\pm 0.30}$ & $0.35_{\pm 0.36}$ & $0.37_{\pm 0.48}$ & $0.43_{\pm 0.09}$ \\
 & Struct & $0.87_{\pm 0.34}$ & $0.15_{\pm 0.36}$ & $0.48_{\pm 0.39}$ & $0.25_{\pm 0.43}$ & $0.46_{\pm 0.07}$ \\
 \midrule
M+T & None & $0.89_{\pm 0.32}$ & $0.23_{\pm 0.42}$ & $0.36_{\pm 0.41}$ & $0.02_{\pm 0.14}$ & $0.48_{\pm 0.10}$ \\
 & Light & $0.86_{\pm 0.35}$ & $0.20_{\pm 0.40}$ & $0.30_{\pm 0.35}$ & $0.05_{\pm 0.22}$ & $0.44_{\pm 0.10}$ \\
 & Struct & $0.85_{\pm 0.36}$ & $0.01_{\pm 0.12}$ & $0.69_{\pm 0.31}$ & $0.01_{\pm 0.09}$ & $0.47_{\pm 0.08}$ \\
  \midrule
D+T & None & $0.86_{\pm 0.34}$ & $0.35_{\pm 0.48}$ & $0.72_{\pm 0.20}$ & $0.05_{\pm 0.21}$ & $0.48_{\pm 0.07}$ \\
 & Light & $0.85_{\pm 0.36}$ & $0.36_{\pm 0.48}$ & $0.72_{\pm 0.20}$ & $0.30_{\pm 0.46}$ & $0.46_{\pm 0.07}$ \\
 & Struct & $0.71_{\pm 0.46}$ & $0.36_{\pm 0.48}$ & $0.83_{\pm 0.17}$ & $0.44_{\pm 0.50}$ & $0.51_{\pm 0.06}$ \\

\bottomrule

\end{tabular}
}
\end{table}

\section{Backbone LLMs}
\label{app:backbone}
Tables~\ref{apptab:8} and~\ref{apptab:9} are the result details of the backbone LLMs variations experiments.

\begin{table}[h]
\centering
\caption{Homogeneous (S1): Overall performance of policies when all backbone LLMs are Qwen3. Report as mean$\pm$std over all agents and all rounds in $Land$ scenario.}
\label{apptab:8}

\resizebox{0.49\textwidth}{!}{
\begin{tabular}{ccccccc}
\toprule
\textbf{Query} & \textbf{Rule} & \textbf{Resp.} & \textbf{Rebuttal} & \textbf{Non-rep.} & \textbf{Evid.} & \textbf{Stance} \\
\midrule
Q1 & None & $0.81_{\pm 0.40}$ & $0.35_{\pm 0.48}$ & $0.18_{\pm 0.35}$ & $0.13_{\pm 0.33}$ & $0.49_{\pm 0.09}$ \\
 & Light & $0.68_{\pm 0.47}$ & $0.28_{\pm 0.45}$ & $0.23_{\pm 0.36}$ & $0.28_{\pm 0.45}$ & $0.44_{\pm 0.08}$ \\
 & Struct & $0.67_{\pm 0.47}$ & $0.04_{\pm 0.20}$ & $0.37_{\pm 0.42}$ & $0.09_{\pm 0.29}$ & $0.47_{\pm 0.07}$ \\
\midrule
Q2 & None & $0.68_{\pm 0.47}$ & $0.25_{\pm 0.43}$ & $0.21_{\pm 0.37}$ & $0.33_{\pm 0.47}$ & $0.47_{\pm 0.08}$ \\
 & Light & $0.64_{\pm 0.48}$ & $0.30_{\pm 0.46}$ & $0.22_{\pm 0.34}$ & $0.25_{\pm 0.43}$ & $0.42_{\pm 0.06}$ \\
 & Struct & $0.68_{\pm 0.47}$ & $0.00_{\pm 0.00}$ & $0.36_{\pm 0.42}$ & $0.01_{\pm 0.12}$ & $0.46_{\pm 0.07}$ \\
\midrule
Q3 & None & $0.76_{\pm 0.43}$ & $0.00_{\pm 0.00}$ & $0.47_{\pm 0.42}$ & $0.00_{\pm 0.00}$ & $0.49_{\pm 0.08}$ \\
 & Light & $0.77_{\pm 0.42}$ & $0.01_{\pm 0.12}$ & $0.42_{\pm 0.40}$ & $0.03_{\pm 0.18}$ & $0.46_{\pm 0.06}$ \\
 & Struct & $0.85_{\pm 0.36}$ & $0.03_{\pm 0.18}$ & $0.52_{\pm 0.39}$ & $0.22_{\pm 0.42}$ & $0.46_{\pm 0.06}$ \\
\midrule
Q4 & None & $0.53_{\pm 0.50}$ & $0.60_{\pm 0.49}$ & $0.16_{\pm 0.34}$ & $0.15_{\pm 0.36}$ & $0.50_{\pm 0.04}$ \\
 & Light & $0.12_{\pm 0.33}$ & $0.56_{\pm 0.50}$ & $0.17_{\pm 0.34}$ & $0.21_{\pm 0.41}$ & $0.51_{\pm 0.07}$ \\
 & Struct & $0.75_{\pm 0.43}$ & $0.39_{\pm 0.49}$ & $0.28_{\pm 0.40}$ & $0.05_{\pm 0.23}$ & $0.50_{\pm 0.04}$ \\
\midrule
Q5 & None & $0.55_{\pm 0.50}$ & $0.20_{\pm 0.40}$ & $0.25_{\pm 0.38}$ & $0.11_{\pm 0.32}$ & $0.44_{\pm 0.05}$ \\
 & Light & $0.75_{\pm 0.43}$ & $0.14_{\pm 0.35}$ & $0.26_{\pm 0.35}$ & $0.01_{\pm 0.12}$ & $0.39_{\pm 0.05}$ \\
 & Struct & $0.73_{\pm 0.44}$ & $0.14_{\pm 0.35}$ & $0.38_{\pm 0.40}$ & $0.09_{\pm 0.29}$ & $0.45_{\pm 0.07}$ \\
\midrule
Overall & None & $0.67_{\pm 0.47}$ & $0.28_{\pm 0.45}$ & $0.25_{\pm 0.39}$ & $0.14_{\pm 0.35}$ & $0.48_{\pm 0.07}$ \\
 & Light & $0.59_{\pm 0.49}$ & $0.26_{\pm 0.44}$ & $0.26_{\pm 0.37}$ & $0.16_{\pm 0.36}$ & $0.44_{\pm 0.08}$ \\
 & Struct & $0.74_{\pm 0.44}$ & $0.12_{\pm 0.33}$ & $0.38_{\pm 0.41}$ & $0.10_{\pm 0.29}$ & $0.47_{\pm 0.07}$ \\
\bottomrule
\end{tabular}
}
\end{table}

\begin{table}[h]
\centering
\caption{Heterogeneous: Overall performance of policies with the variant of the backbones. Report as mean$\pm$std over all agents and all rounds in $Land$ scenario. }
\label{apptab:9}
\resizebox{0.49\textwidth}{!}{
\begin{tabular}{ccccccc}
\toprule
\textbf{Query} & \textbf{Rule} & \textbf{Resp.} & \textbf{Rebuttal} & \textbf{Non-rep.} & \textbf{Evid.} & \textbf{Stance} \\
\midrule
Q1 & None & $0.75_{\pm 0.44}$ & $0.35_{\pm 0.48}$ & $0.37_{\pm 0.40}$ & $0.07_{\pm 0.25}$ & $0.50_{\pm 0.10}$ \\
 & Light & $0.81_{\pm 0.40}$ & $0.31_{\pm 0.46}$ & $0.37_{\pm 0.35}$ & $0.49_{\pm 0.50}$ & $0.44_{\pm 0.10}$ \\
 & Struct & $0.79_{\pm 0.41}$ & $0.12_{\pm 0.33}$ & $0.61_{\pm 0.37}$ & $0.23_{\pm 0.42}$ & $0.55_{\pm 0.07}$ \\
\midrule
Q2 & None & $0.70_{\pm 0.46}$ & $0.31_{\pm 0.46}$ & $0.36_{\pm 0.39}$ & $0.07_{\pm 0.26}$ & $0.46_{\pm 0.09}$ \\
 & Light & $0.69_{\pm 0.47}$ & $0.28_{\pm 0.45}$ & $0.26_{\pm 0.32}$ & $0.33_{\pm 0.47}$ & $0.46_{\pm 0.08}$ \\
 & Struct & $0.67_{\pm 0.47}$ & $0.11_{\pm 0.31}$ & $0.61_{\pm 0.34}$ & $0.28_{\pm 0.45}$ & $0.48_{\pm 0.09}$ \\
\midrule
Q3 & None & $0.85_{\pm 0.35}$ & $0.01_{\pm 0.08}$ & $0.46_{\pm 0.40}$ & $0.01_{\pm 0.08}$ & $0.48_{\pm 0.08}$ \\
 & Light & $0.85_{\pm 0.35}$ & $0.00_{\pm 0.00}$ & $0.37_{\pm 0.37}$ & $0.24_{\pm 0.43}$ & $0.49_{\pm 0.07}$ \\
 & Struct & $0.72_{\pm 0.45}$ & $0.05_{\pm 0.21}$ & $0.73_{\pm 0.30}$ & $0.25_{\pm 0.43}$ & $0.47_{\pm 0.09}$ \\
\midrule
Q4 & None & $0.57_{\pm 0.50}$ & $0.47_{\pm 0.50}$ & $0.40_{\pm 0.39}$ & $0.10_{\pm 0.30}$ & $0.51_{\pm 0.08}$ \\
 & Light & $0.49_{\pm 0.50}$ & $0.47_{\pm 0.50}$ & $0.30_{\pm 0.36}$ & $0.65_{\pm 0.48}$ & $0.51_{\pm 0.07}$ \\
 & Struct & $0.59_{\pm 0.49}$ & $0.34_{\pm 0.47}$ & $0.52_{\pm 0.38}$ & $0.19_{\pm 0.40}$ & $0.52_{\pm 0.09}$ \\
\midrule
Q5 & None & $0.69_{\pm 0.47}$ & $0.17_{\pm 0.38}$ & $0.35_{\pm 0.39}$ & $0.04_{\pm 0.20}$ & $0.45_{\pm 0.08}$ \\
 & Light & $0.77_{\pm 0.42}$ & $0.43_{\pm 0.50}$ & $0.36_{\pm 0.36}$ & $0.40_{\pm 0.49}$ & $0.40_{\pm 0.08}$ \\
 & Struct & $0.71_{\pm 0.46}$ & $0.19_{\pm 0.40}$ & $0.58_{\pm 0.37}$ & $0.16_{\pm 0.37}$ & $0.47_{\pm 0.08}$ \\
\midrule
Overall & None & $0.71_{\pm 0.45}$ & $0.26_{\pm 0.44}$ & $0.39_{\pm 0.40}$ & $0.06_{\pm 0.23}$ & $0.48_{\pm 0.09}$ \\
 & Light & $0.72_{\pm 0.45}$ & $0.30_{\pm 0.46}$ & $0.33_{\pm 0.35}$ & $0.42_{\pm 0.49}$ & $0.46_{\pm 0.09}$ \\
 & Struct & $0.70_{\pm 0.46}$ & $0.16_{\pm 0.37}$ & $0.61_{\pm 0.36}$ & $0.22_{\pm 0.42}$ & $0.50_{\pm 0.09}$ \\

\bottomrule

\end{tabular}
}
\end{table}

\section{Agent-level performance}
\label{app:more}

To further evaluate the effectiveness of the policy-parameterized framework in a more-agent system, we added two more agents (Farmer 2 and 3) driven by  Llama3 and Mistral, respectively.  Farmer 1 and Farmer 2 are designed to share similar personas, whereas Farmer 3 is assigned a different behavioural role to introduce diversity. Three Farmer agents share a knowledge base. The results in Table~\ref{tab:more_agent} show that the policy-parameterized prompts remain effective: each agent still shows differences across various metrics under different rules. Table~\ref{tab:agent_performance} shows the overall performance of policies of each agent under different Backbone LLMs.

\begin{table}[h]
\centering
\caption{Performance of policies of each agent after adding two Farmer agents.}
\label{tab:more_agent}

\resizebox{0.49\textwidth}{!}{
\begin{tabular}{cccccccc}
\toprule
 \textbf{Rule} & \textbf{Agent} & \textbf{Resp.} & \textbf{Rebuttal} & \textbf{Non-rep.} & \textbf{Evid.} & \textbf{Stance} \\
\midrule
None  & Comm. & $0.90 \pm 0.00$ & $0.08 \pm 0.04$ & $0.84 \pm 0.03$ & $0.28 \pm 0.14$ & $0.46 \pm 0.05$ \\
 & Conser.& $0.87 \pm 0.03$ & $0.24 \pm 0.17$ & $0.71 \pm 0.06$ & $0.02 \pm 0.03$ & $0.49 \pm 0.06$ \\
 & Farmer 1(Qwen3) & $0.88 \pm 0.02$ & $0.50 \pm 0.33$ & $0.52 \pm 0.15$ & $0.13 \pm 0.13$ & $0.59 \pm 0.04$ \\
 & Farmer 2(Llama)  & $0.87 \pm 0.03$ & $0.63 \pm 0.21$ & $0.59 \pm 0.16$ & $0.10 \pm 0.10$ & $0.68 \pm 0.06$ \\
 & Farmer 3(Mistral)   & $0.88 \pm 0.02$ & $0.26 \pm 0.17$ & $0.80 \pm 0.08$ & $0.33 \pm 0.20$ & $0.51 \pm 0.02$ \\
 \midrule
Light & Comm.& $0.90 \pm 0.01$ & $0.09 \pm 0.07$ & $0.63 \pm 0.03$ & $0.21 \pm 0.14$ & $0.40 \pm 0.03$ \\
 & Conser. & $0.88 \pm 0.02$ & $0.21 \pm 0.21$ & $0.71 \pm 0.07$ & $0.17 \pm 0.04$ & $0.44 \pm 0.06$ \\
 & Farmer 1(Qwen3) & $0.86 \pm 0.04$ & $0.58 \pm 0.28$ & $0.39 \pm 0.06$ & $0.58 \pm 0.17$ & $0.55 \pm 0.05$ \\
 & Farmer 2(Llama) & $0.87 \pm 0.05$ & $0.62 \pm 0.25$ & $0.49 \pm 0.09$ & $0.64 \pm 0.17$ & $0.62 \pm 0.05$ \\
 & Farmer 3(Mistral)  & $0.89 \pm 0.02$ & $0.17 \pm 0.10$ & $0.65 \pm 0.04$ & $0.54 \pm 0.08$ & $0.49 \pm 0.03$ \\
\midrule
Struct & Comm. & $0.90 \pm 0.00$ & $0.10 \pm 0.04$ & $0.83 \pm 0.03$ & $0.11 \pm 0.05$ & $0.42 \pm 0.02$ \\
 & Conser. & $0.78 \pm 0.03$ & $0.19 \pm 0.07$ & $0.78 \pm 0.02$ & $0.26 \pm 0.08$ & $0.48 \pm 0.01$ \\
 & Farmer 1(Qwen3) & $0.78 \pm 0.06$ & $0.23 \pm 0.13$ & $0.74 \pm 0.04$ & $0.33 \pm 0.14$ & $0.55 \pm 0.03$ \\
 & Farmer 2(Llama)  & $0.80 \pm 0.04$ & $0.25 \pm 0.16$ & $0.75 \pm 0.05$ & $0.40 \pm 0.10$ & $0.61 \pm 0.03$ \\
 & Farmer 3(Mistral) & $0.89 \pm 0.02$ & $0.27 \pm 0.17$ & $0.80 \pm 0.05$ & $0.34 \pm 0.10$ & $0.48 \pm 0.02$ \\

\bottomrule
\end{tabular}
}
\end{table}

\begin{table}[h]
\centering
\caption{Overall performance of policies of each agent under different Backbone LLMs}
\label{tab:agent_performance}

\resizebox{0.49\textwidth}{!}{
\begin{tabular}{ccccccc}
\toprule
\textbf{Rule} & \textbf{Agent} & \textbf{Resp.} & \textbf{Rebuttal} & \textbf{Non-rep.} & \textbf{Evid.} & \textbf{Stance} \\
\midrule
 None & Comm.(mistral)  & $0.48 \pm 0.20$ & $0.02 \pm 0.04$ & $0.22 \pm 0.10$ & $0.14 \pm 0.13$ & $0.41 \pm 0.03$ \\
& Conser.(Llama)  & $0.72 \pm 0.28$ & $0.23 \pm 0.34$ & $0.30 \pm 0.19$ & $0.26 \pm 0.26$ & $0.48 \pm 0.04$ \\
 & Farmer(Qwen) & $0.80 \pm 0.12$ & $0.58 \pm 0.35$ & $0.24 \pm 0.06$ & $0.04 \pm 0.08$ & $0.54 \pm 0.03$ \\

\midrule
 Light & Comm.(mistral) & $0.40 \pm 0.16$ & $0.04 \pm 0.07$ & $0.22 \pm 0.05$ & $0.23 \pm 0.21$ & $0.39 \pm 0.04$ \\
 & Conser.(Llama) & $0.65 \pm 0.31$ & $0.21 \pm 0.29$ & $0.31 \pm 0.13$ & $0.13 \pm 0.14$ & $0.44 \pm 0.04$ \\
& Farmer(Qwen) & $0.73 \pm 0.29$ & $0.53 \pm 0.38$ & $0.26 \pm 0.08$ & $0.11 \pm 0.12$ & $0.51 \pm 0.05$ \\

 \midrule
 Struct & Comm.(mistral) &  $0.57 \pm 0.12$ & $0.03 \pm 0.06$ & $0.37 \pm 0.07$ & $0.17 \pm 0.24$ & $0.41 \pm 0.04$ \\
 & Conser.(Llama) &  $0.78 \pm 0.07$ & $0.20 \pm 0.28$ & $0.43 \pm 0.10$ & $0.11 \pm 0.09$ & $0.48 \pm 0.02$ \\
 & Farmer(Qwen) & $0.86 \pm 0.02$ & $0.13 \pm 0.15$ & $0.35 \pm 0.09$ & $0.00 \pm 0.01$ & $0.51 \pm 0.03$ \\

  \midrule
 None & Comm.(Qwen) &  $0.52 \pm 0.19$ & $0.01 \pm 0.02$ & $0.25 \pm 0.14$ & $0.03 \pm 0.06$ & $0.42 \pm 0.03$  \\
 & Conser.(Mistral) & $0.83 \pm 0.14$ & $0.19 \pm 0.20$ & $0.53 \pm 0.06$ & $0.07 \pm 0.08$ & $0.45 \pm 0.03$ \\
& Farmer(Llama) & $0.79 \pm 0.17$ & $0.58 \pm 0.36$ & $0.38 \pm 0.08$ & $0.08 \pm 0.11$ & $0.57 \pm 0.03$ \\

\midrule
 Light & Comm.(Qwen) & $0.59 \pm 0.18$ & $0.02 \pm 0.05$ & $0.22 \pm 0.09$ & $0.31 \pm 0.26$ & $0.41 \pm 0.04$  \\
& Conser.(Mistral) & $0.87 \pm 0.05$ & $0.20 \pm 0.24$ & $0.45 \pm 0.05$ & $0.21 \pm 0.14$ & $0.42 \pm 0.04$ \\
 & Farmer(Llama) & $0.71 \pm 0.30$ & $0.66 \pm 0.34$ & $0.33 \pm 0.06$ & $0.74 \pm 0.06$ & $0.55 \pm 0.04$ \\

 \midrule
 Struct & Comm.(Qwen) &  $0.61 \pm 0.06$ & $0.06 \pm 0.07$ & $0.45 \pm 0.09$ & $0.24 \pm 0.12$ & $0.44 \pm 0.04$  \\
 & Conser.(Mistral) & $0.86 \pm 0.04$ & $0.22 \pm 0.20$ & $0.74 \pm 0.10$ & $0.01 \pm 0.02$ & $0.48 \pm 0.03$ \\
 & Farmer(Llama) & $0.62 \pm 0.13$ & $0.21 \pm 0.11$ & $0.64 \pm 0.06$ & $0.41 \pm 0.08$ & $0.58 \pm 0.02$  \\
 \midrule
None & Comm.(Llama) & $0.75 \pm 0.12$ & $0.03 \pm 0.05$ & $0.38 \pm 0.04$ & $0.13 \pm 0.08$ & $0.44 \pm 0.03$ \\
 & Conser.(Qwen) & $0.76 \pm 0.13$ & $0.34 \pm 0.35$ & $0.33 \pm 0.19$ & $0.10 \pm 0.12$ & $0.46 \pm 0.06$ \\
 & Farmer (Mistral) & $0.85 \pm 0.07$ & $0.31 \pm 0.28$ & $0.65 \pm 0.08$ & $0.12 \pm 0.08$ & $0.52 \pm 0.03$ \\  \midrule
 Light & Comm.(Llama) & $0.89 \pm 0.01$ & $0.05 \pm 0.09$ & $0.32 \pm 0.07$ & $0.25 \pm 0.18$ & $0.42 \pm 0.02$ \\
 & Conser.(Qwen) & $0.76 \pm 0.26$ & $0.37 \pm 0.37$ & $0.28 \pm 0.12$ & $0.04 \pm 0.04$ & $0.44 \pm 0.05$ \\
 & Farmer (Mistral) & $0.90 \pm 0.01$ & $0.30 \pm 0.26$ & $0.46 \pm 0.13$ & $0.42 \pm 0.23$ & $0.51 \pm 0.03$ \\ \midrule
 Struct & Comm.(Llama) & $0.88 \pm 0.02$ & $0.00 \pm 0.01$ & $0.63 \pm 0.04$ & $0.35 \pm 0.20$ & $0.47 \pm 0.02$ \\
 & Conser.(Qwen) & $0.78 \pm 0.03$ & $0.12 \pm 0.09$ & $0.59 \pm 0.05$ & $0.08 \pm 0.10$ & $0.49 \pm 0.01$ \\
 & Farmer (Mistral) & $0.87 \pm 0.05$ & $0.12 \pm 0.08$ & $0.78 \pm 0.06$ & $0.29 \pm 0.11$ & $0.53 \pm 0.01$ \\

\bottomrule
\end{tabular}
}
\end{table}

\section{Adaptive Weight}
\label{app:adaptive}

Table~\ref{apptab:10} shows the experitment details of the effectiveness of adaptive weights.

\begin{table}[H]
\centering
\caption{Overall performance with adaptive weight setting. Report as mean$\pm$std over all agents and all rounds in $Land$ scenario. }
\label{apptab:10}

\resizebox{0.49\textwidth}{!}{
\begin{tabular}{ccccccc}
\toprule
\textbf{Query} & \textbf{Rule} & \textbf{Resp.} & \textbf{Rebuttal} & \textbf{Non-rep.} & \textbf{Evid.} & \textbf{Stance} \\
\midrule
Q1 & None & $0.88_{\pm 0.33}$ & $0.21_{\pm 0.41}$ & $0.50_{\pm 0.38}$ & $0.23_{\pm 0.42}$ & $0.53_{\pm 0.07}$ \\
 & Light & $0.89_{\pm 0.32}$ & $0.51_{\pm 0.50}$ & $0.36_{\pm 0.35}$ & $0.47_{\pm 0.50}$ & $0.50_{\pm 0.07}$ \\
 & Struct & $0.83_{\pm 0.38}$ & $0.17_{\pm 0.38}$ & $0.54_{\pm 0.40}$ & $0.15_{\pm 0.36}$ & $0.51_{\pm 0.07}$ \\
\midrule
Q2 & None & $0.87_{\pm 0.33}$ & $0.38_{\pm 0.49}$ & $0.49_{\pm 0.39}$ & $0.19_{\pm 0.40}$ & $0.48_{\pm 0.07}$ \\
 & Light & $0.89_{\pm 0.31}$ & $0.39_{\pm 0.49}$ & $0.39_{\pm 0.33}$ & $0.17_{\pm 0.38}$ & $0.44_{\pm 0.05}$ \\
 & Struct & $0.81_{\pm 0.40}$ & $0.11_{\pm 0.32}$ & $0.49_{\pm 0.39}$ & $0.17_{\pm 0.38}$ & $0.47_{\pm 0.07}$ \\
\midrule
Q3 & None & $0.90_{\pm 0.30}$ & $0.00_{\pm 0.00}$ & $0.39_{\pm 0.39}$ & $0.12_{\pm 0.33}$ & $0.45_{\pm 0.06}$ \\
 & Light & $0.90_{\pm 0.30}$ & $0.07_{\pm 0.26}$ & $0.43_{\pm 0.36}$ & $0.27_{\pm 0.44}$ & $0.44_{\pm 0.06}$ \\
 & Struct & $0.89_{\pm 0.31}$ & $0.02_{\pm 0.14}$ & $0.54_{\pm 0.40}$ & $0.10_{\pm 0.30}$ & $0.45_{\pm 0.06}$ \\
\midrule
Q4 & None & $0.79_{\pm 0.41}$ & $0.41_{\pm 0.49}$ & $0.35_{\pm 0.37}$ & $0.17_{\pm 0.38}$ & $0.49_{\pm 0.05}$ \\
 & Light & $0.70_{\pm 0.46}$ & $0.55_{\pm 0.50}$ & $0.33_{\pm 0.35}$ & $0.42_{\pm 0.49}$ & $0.49_{\pm 0.08}$ \\
 & Struct & $0.79_{\pm 0.41}$ & $0.18_{\pm 0.39}$ & $0.47_{\pm 0.40}$ & $0.24_{\pm 0.43}$ & $0.51_{\pm 0.07}$ \\
\midrule
Q5 & None & $0.86_{\pm 0.35}$ & $0.31_{\pm 0.47}$ & $0.51_{\pm 0.38}$ & $0.17_{\pm 0.37}$ & $0.46_{\pm 0.06}$ \\
 & Light & $0.87_{\pm 0.34}$ & $0.45_{\pm 0.50}$ & $0.35_{\pm 0.34}$ & $0.13_{\pm 0.34}$ & $0.41_{\pm 0.05}$ \\
 & Struct & $0.79_{\pm 0.41}$ & $0.18_{\pm 0.39}$ & $0.55_{\pm 0.40}$ & $0.15_{\pm 0.35}$ & $0.47_{\pm 0.06}$ \\

\midrule
Overall & None & $0.86_{\pm 0.35}$ & $0.26_{\pm 0.44}$ & $0.45_{\pm 0.39}$ & $0.18_{\pm 0.38}$ & $0.48_{\pm 0.07}$ \\
 & Light & $0.85_{\pm 0.36}$ & $0.39_{\pm 0.49}$ & $0.37_{\pm 0.35}$ & $0.29_{\pm 0.46}$ & $0.45_{\pm 0.07}$ \\
 & Struct & $0.82_{\pm 0.38}$ & $0.13_{\pm 0.34}$ & $0.52_{\pm 0.40}$ & $0.16_{\pm 0.37}$ & $0.48_{\pm 0.07}$ \\
\bottomrule

\end{tabular}
}
\end{table}

\section{Ablation Study on EDU}
\label{app:ablation_edu}
Table~\ref{tab:query_rule_ablation} shows the ablation study results on EDU scenario.

\begin{table}[t]
\centering
\caption{Ablation Study on EDU scenario}
\label{tab:query_rule_ablation}
\resizebox{0.49\textwidth}{!}{
\begin{tabular}{c c | c c c c c}
\toprule
Query & Rule & Resp. & Reb. & Non-rep. & Evid. & Stance \\
\midrule
M+D & None   & $0.89\pm0.32$ & $0.02\pm0.15$ & $0.42\pm0.39$ & $0.06\pm0.23$ & $0.46\pm0.08$ \\
    & Light  & $0.89\pm0.32$ & $0.01\pm0.12$ & $0.39\pm0.36$ & $0.14\pm0.35$ & $0.45\pm0.09$ \\
    & Struct & $0.89\pm0.31$ & $0.02\pm0.15$ & $0.46\pm0.38$ & $0.05\pm0.22$ & $0.45\pm0.08$ \\
\midrule
T+D & None   & $0.85\pm0.35$ & $0.46\pm0.50$ & $0.74\pm0.24$ & $0.13\pm0.34$ & $0.50\pm0.08$ \\
    & Light  & $0.85\pm0.35$ & $0.34\pm0.47$ & $0.69\pm0.25$ & $0.55\pm0.50$ & $0.48\pm0.09$ \\
    & Struct & $0.80\pm0.40$ & $0.30\pm0.46$ & $0.80\pm0.17$ & $0.31\pm0.46$ & $0.53\pm0.08$ \\
\midrule
T+M & None   & $0.86\pm0.35$ & $0.17\pm0.38$ & $0.45\pm0.39$ & $0.09\pm0.29$ & $0.46\pm0.10$ \\
    & Light  & $0.89\pm0.31$ & $0.10\pm0.30$ & $0.37\pm0.35$ & $0.08\pm0.27$ & $0.44\pm0.09$ \\
    & Struct & $0.88\pm0.33$ & $0.04\pm0.20$ & $0.65\pm0.30$ & $0.01\pm0.11$ & $0.46\pm0.08$ \\
\bottomrule
\end{tabular}
}
\end{table}

\end{document}